\documentclass{article} 
\usepackage{iclr2020_conference,times}

\usepackage{amsmath,amsfonts,bm}

\def\eqref#1{equation~\ref{#1}}

\def\1{\bm{1}}

\DeclareMathAlphabet{\mathsfit}{\encodingdefault}{\sfdefault}{m}{sl}
\SetMathAlphabet{\mathsfit}{bold}{\encodingdefault}{\sfdefault}{bx}{n}

\usepackage[utf8]{inputenc} 
\usepackage[T1]{fontenc}    
\usepackage{hyperref}       
\usepackage{url}            
\usepackage{booktabs}       
\usepackage{amsfonts}       
\usepackage{nicefrac}       
\usepackage{microtype}      
\usepackage{amsmath}
\usepackage{graphicx}
\usepackage{multirow}
\usepackage{algorithm}
\usepackage{algorithmic}
\usepackage{tabularx}
\usepackage{subfigure}
\usepackage{amsmath}
\usepackage{enumitem}
\usepackage{color}
\usepackage{float}
\newboolean{Appendix}
\setboolean{Appendix}{true}
\newcolumntype{Y}{>{\centering\arraybackslash}X}
\newcolumntype{b}{>{\hsize=8\hsize}X}
\newcommand{\GenAttack}{TREMBA}
\newcommand{\TGenAttack}{TREMBA$_{\text{OSP}}$ }
\title{Black-Box Adversarial Attack with Transferable Model-based Embedding}

\author{Zhichao Huang, Tong Zhang  \\
The Hong Kong University of Science and Technology\\
\texttt{zhuangbx@connect.ust.hk, tongzhang@tongzhang-ml.org}
}

\iclrfinalcopy 
\begin{document}

\maketitle

\begin{abstract}
We present a new method for black-box adversarial attack. Unlike previous methods that combined transfer-based and scored-based methods by using the gradient or initialization of a surrogate white-box model, this new method tries to learn a low-dimensional embedding using a pretrained model, and then performs efficient search within the embedding space to attack an unknown target network. The method produces adversarial perturbations with high level semantic patterns that are easily transferable. We show that this approach can greatly improve the query efficiency of black-box adversarial attack across different target network architectures. We evaluate our approach on MNIST, ImageNet and Google Cloud Vision API, resulting in a significant reduction on the number of queries. We also attack adversarially defended networks on CIFAR10 and ImageNet, where our method not only reduces the number of queries, but also improves the attack success rate.
\end{abstract}
\section{Introduction}

The wide adoption of neural network models in modern applications has caused major security concerns, as such models are known to be vulnerable to adversarial examples that can fool neural networks to make wrong predictions \citep{adversarial}. Methods to attack neural networks can be divided into two categories based on whether the parameters of the neural network are assumed to be known to the attacker: white-box attack and black-box attack. There are several approaches to find adversarial examples for black-box neural networks. The transfer-based attack methods first pretrain a source model and then generate adversarial examples using a standard white-box attack method on the source model to attack an unknown target network \citep{goodfellow2014explaining, madry, cw, transfer}. The score-based attack requires a loss-oracle, which enables the attacker to query the target network at multiple points to approximate its gradient. The attacker can then apply the white-box attack techniques with the approximated gradient \citep{zoo, nes, autozoom}. 

A major problem of the transfer-based attack is that it can not achieve very high success rate. And transfer-based attack is weak in targeted attack. On the contrary, the success rate of score-based attack has only small gap to the white-box attack but it requires many queries. Thus, it is natural to combine the two black-box attack approaches, so that we can take advantage of a pretrained white-box source neural network to perform more efficient search to attack an unknown target black-box model. 

In fact, in the recent NeurIPS 2018 Adversarial Vision Challenge \citep{adversarial_vision_challenge}, many teams transferred adversarial examples from a source network as the starting point to carry out black-box boundary attack \citep{boundary}. $\mathcal{N}$Attack also used a regression network as initialization in the score-based attack \citep{nattack}. The transferred adversarial example could be a good starting point that lies close to the decision boundary for the target network and accelerate further optimization. P-RGF \citep{prgf} used the gradient information from the source model to accelerate searching process. However, gradient information is localized and sometimes it is misleading. In this paper, we push the idea of using a pretrained white-box source network to guide black-box attack significantly further, by proposing a method called TRansferable EMbedding based Black-box Attack (\GenAttack). {\GenAttack} contains two stages: (1) train an encoder-decoder that can effectively generate adversarial perturbations for the source network with a low-dimensional embedding space; (2) apply NES (Natural Evolution Strategy) of \citep{wierstra2014natural} to the low-dimensional embedding space of the pretrained generator to search adversarial examples for the target network. {\GenAttack} uses global information of the source model, capturing high level semantic adversarial features that are insensitive to different models. Unlike noise-like perturbations, such perturbations would have much higher transferablity across different models. Therefore we could gain query efficiency by performing queries in the embedding space. 

We note that there have been a number of earlier works on using generators to produce adversarial perturbations in the white-box setting \citep{atn, advgan, wang2018a}. While black-box attacks were also considered there, they focused on training generators with dynamic distillation. These early approaches required many queries to fine-tune the classifier for different target networks, which may not be practical for real applications. While our approach also relies on a generator, we train it as an encoder-decoder that produces a low-dimensional embedding space. By applying a standard black-box attack method such as NES on the embedding space, adversarial perturbations can be found efficiently for a target model.

It is worth noting that the embedding approach has also been used in AutoZOOM \citep{autozoom}. However, it only trained the autoencoder to reconstruct the input, and it did not take advantage of the information of a pretrained network. Although it also produces structural perturbations, these perturbations are usually not suitable for attacking regular networks and sometimes its performance is even worse than directly applying NES to the images \citep{prgf, guo2019simple}. {\GenAttack}, on the other hand, tries to learn an embedding space that can efficiently generate adversarial perturbations for a pretrained source network. Compared to AutoZOOM, our new method produces adversarial perturbation with high level semantic features that could hugely affect arbitrary target networks, resulting in significantly lower number of queries.

We summarize our contributions as follows:
\begin{enumerate}
    \item We propose {\GenAttack}, an attack method that explores a novel way to utilize the information of a pretrained source network to improve the query efficiency of black-box attack on a target network. 
    \item We show that {\GenAttack} can produce adversarial perturbations with high level semantic patterns, which are effective across different networks, resulting in much lower queries on MNIST and ImageNet especially for the targeted attack that has low transferablity.
    \item We demonstrate that {\GenAttack} can be applied to SOTA defended models \citep{madry, denosing}. Compared with other black-box attacks, {\GenAttack} increases success rate by approximately $10\%$ while reduces the number of queries by more than $50\%$. 
\end{enumerate}

\section{Related Works}

There have been a vast literature on adversarial examples. We will cover the most relevant topics including white-box attack, black-box attack and defense methods. 

\textbf{White-Box Attack} White-box attack requires the full knowledge of the target model. It was first discovered by \citep{adversarial} that adversarial examples could be found by solving an optimization problem with L-BFGS \citep{lbgfs}. Later on, other methods were proposed to find adversarial examples with improved success rate and efficiency \citep{goodfellow2014explaining, pgd, jsma, deepfool}. More recently, it was shown that generators can also construct adversarial noises with high success rate \citep{advgan, atn}.

\textbf{Black-Box Attack} Black-box attack can be divided into three categories: transfer-based, score-based and decision-based. It is well known that adversaries have high transferablity across different networks \citep{transfer}. Transfer-based methods generate adversarial noises on a source model and then transfer it to an unknown target network. It is known that targeted attack is harder than un-targeted attack for transfer-based methods, and using an ensemble of source models can improve the success rate \citep{ensembleattack}. Score-based attack assumes that the attacker can query the output scores of the target network. The attacker usually uses sampling methods to approximate the true gradient \citep{zoo, nes, nattack,fw}. AutoZOOM tried to improve the query efficiency by reducing the sampling space with a bilinear transformation or an autoencoder \citep{autozoom}. \citep{bandit} incorporated data and time prior to accelerate attacking. In contrast to the gradient based method, \citep{moon2019parsimonious} used combinatorial optimization to achieve good efficiency. In decision-based attack, the attacker only knows the output label of the classifier. Boundary attack and its variants are very powerful in this setting \citep{boundary, CMAboundary}. In NeutIPS 2018 Adversarial Vision Challenge \citep{adversarial_vision_challenge}, some teams combined transfer-based attack and decision-based attack in their attacking methods \citep{brunner2018guessing}. And in a similar spirit, $\mathcal{N}$Attack also used a regression network as initialization in score-based attack \citep{nattack}. Gradient information from the surrogate model could also be used to accelerate the scored-based attack \citep{prgf} . 

\textbf{Defense Methods} Several methods have been proposed to overcome the vulnerability of neural networks. Gradient masking based methods add non-differential operations in the model, interrupting the backward pass of gradients. However, they are vulnerable to adversarial attacks with the approximated gradient \citep{BPDA, nattack}. Adversarial training is the SOTA method that can be used to improve the robustness of neural networks. Adversarial training is a minimax game. The outside minimizer performs regular training of the neural network, and the inner maximizer finds a perturbation of the input to attack the network. The inner maximization process can be approximated with FGSM \citep{goodfellow2014explaining}, PGD \citep{madry}, adversarial generator \citep{wang2018a} etc. Moreover, feature denoising can improve the robustness of neural networks on ImageNet \citep{denosing}.

\section{Black-Box Adversarial Attack with Generator}\label{sec:method}
Consider a DNN classifier $F(x)$. Let $x \in [0,1]^{\text{dim}(x)}$ be an input, and let $F(x)$ be the output vector obtained before the softmax layer. We denote $F(x)_i$ as the $i$-th component for the output vector and $y$ as the label for the input. For un-targeted attack, our goal is to find a small perturbation $\delta$ such that the classifier predicts the wrong label, i.e. $\mathop{\arg\max} \ F(x + \delta) \neq y$. And for targeted attack, we want the classifier to predicts the target label $t$, i.e. $\mathop{\arg\max} \ F(x + \delta) = t$. The perturbation $\delta$ is usually bounded by $\ell_p$ norm: $\|\delta\|_p \le \varepsilon$, with a small $\varepsilon>0$.

Adversarial perturbations often have high transferablity across different DNNs. Given a white-box source DNN $F_s$ with known architecture and parameters, we can transfer its white-box adversarial perturbation $\delta_s$ to a black-box target DNN $F_t$ with reasonably good success rate. It is known that even if $x+\delta_s$ fails to be an adversarial example, $\delta_s$ can still act as a good starting point for searching adversarial examples using a score-based attack method. This paper shows that the information of $F_s$ can be further utilized to train a generator, and performing search on its embedding space leads to more efficient black-box attacks of an unknown target network $F_t$.

\subsection{Generating Adversarial Perturbations with Generator}

Adversarial perturbations can be generated by a generator network $\mathcal{G}$. We explicitly divide the generator into two parts: an encoder $\mathcal{E}$ and a decoder $\mathcal{D}$. The encoder takes the origin input $x$ and output a latent vector $z = \mathcal{E}(x)$, where $\text{dim}(z) \ll \text{dim}(x)$. The decoder takes $z$ as the input and outputs an adversarial perturbation $\delta = \varepsilon\tanh (\mathcal{D}(z))$ with $\text{dim}(\delta) = \text{dim}(x)$. 
In our new method, we will train the generator $\mathcal{G}$ so that $\delta = \varepsilon \tanh(\mathcal{G}(x))$ can fool the source network $F_s$.

Suppose we have a training set $\left\{\left(x_{1}, y_{1}\right), \ldots,\left(x_{n}, y_{n}\right)\right\}$, where $x_i$ denotes the input and $y_i$ denotes its label. For un-targeted attack, we train the desired generator by minimizing the hinge loss used in the C\&W attack \citep{cw}:
\begin{equation}
    \mathcal{L}_\text{untarget}(x_i,y_i) = \max \left(F_s(\varepsilon \tanh(\mathcal{G}(x_i))+x_i)_{y_i}-\max _{j \neq y_i} F_s(\varepsilon \tanh(\mathcal{G}(x_i))+x_i)_{j},-\kappa\right) ,
    \label{eqn:min}
\end{equation}
And for targeted, we use
\begin{equation}
    \mathcal{L}_\text{target}(x_i, t) =\max \left(\max _{j \neq t} F_s(\varepsilon \tanh(\mathcal{G}(x_i))+x_i)_{j}-F_s(\varepsilon \tanh(\mathcal{G}(x_i))+x_i)_{t},-\kappa\right) ,
    \label{eqn:min-target}
\end{equation}
where $t$ denotes the targeted class and $\kappa$ is the margin parameter that can be used to adjust transferability of the generator. A higher value of $\kappa$ leads to higher transferability to other models \citep{cw}. We focus on $\ell_\infty$ norm in this work. By adding point-wise $\tanh$ function to an unnormalized output $\mathcal{D}(z)$, and scaling it with $\varepsilon$, $\delta=\varepsilon \tanh(\mathcal{D}(z))$ is already bounded as $\|\delta\|_\infty < \varepsilon$. Therefore we employ this transformation, so that we do not need to impose the infinity norm constraint explicitly. While hinge loss is employed in this paper, we believe other loss functions such the cross entropy loss will also work.

\subsection{Search over latent space with NES}

Given a new black-box DNN classifier $F_t(x)$, for which we can only query its output at any given point $x$. As in \citep{nes, wierstra2014natural}, we can employ NES to approximate the gradient of a properly defined surrogate loss in order to find an adversarial example. Denote the surrogate loss by $\mathcal{L}$, rather than calculating $\nabla_{\delta} \mathcal{L}(x+\delta, y)$ directly, NES update $\delta$ by using $ \nabla_{\delta}\mathbb{E}_{\omega \sim \mathcal{N}(\delta, \sigma^2)}[L(x+\omega, y)]$, which can be transformed into $\mathbb{E}_{\omega \sim \mathcal{N}(\delta, \sigma^2)}[L(x+\omega, y) \nabla_{\omega} \log(\mathcal{N}(\omega|\delta, \sigma^2))]$. The expectation can be approximated by taking finite samples. And we could use the following equation to iteratively update $\delta$:
\begin{equation}
    \delta_{t+1} = \prod_{[-\varepsilon, \varepsilon]}(\delta_t - \eta\cdot \text{sign}(\frac{1}{b}\sum_{k=1}^b \mathcal{L}(x+\omega_k, y)\nabla \log \mathcal{N}(\omega_k|\delta_{t}, \sigma^2))) ,
    \label{eqn: nes_image}
\end{equation}
where $\eta$ is the learning rate, $b$ is the minibatch sample size, $\omega_k$ is the sample from the gaussian distribution and $\prod_{[-\varepsilon, \varepsilon]}$ represents a clipping operation, which projects $\delta$ onto the $\ell_\infty$ ball. The sign function provides an approximation of the gradient, which has been widely used in adversarial attack \citep{nes, madry}. However, it is observed that more effective attacks can be obtained by removing the sign function \citep{nattack-iclr}. Therefore in this work, we remove the sign function from Eqn (\ref{eqn: nes_image}) and directly use the estimated gradient. 

Instead of performing search on the input space, {\GenAttack} performs search on the embedding space $z$. The generator $\mathcal{G}$ explores the weakness of the source DNN $F_s$ so that $\mathcal{D}$ produces perturbations that can effective attack $F_s$. For a different unknown target network $F_t$, we show that our method can still generate perturbations leading to more effective attack of $F_t$. Given an input $x$ and its label $y$, we choose a starting point $z^0 = \mathcal{E}(x)$. The gradient of $z^t$ given by NES can be estimated as:
\begin{align}
    \nabla_{z^t} \mathcal{L}(x+\varepsilon\tanh(\mathcal{D}(z^t)), y) &\approx \nabla_{z^t} \mathbb{E}_{\nu \sim \mathcal{N}(z^t,\sigma^2)} \left[\mathcal{L}(x+\varepsilon\tanh(\mathcal{D}(\nu)), y)\right] \\\nonumber
    &\approx \frac{1}{b} \sum_{k=1}^b \mathcal{L}(x+\varepsilon\tanh(\mathcal{D}(\nu_k)),y) \nabla_{z^t} \log \mathcal{N}(\nu_k|z^t,\sigma^2) .
\end{align}
where $\nu_k$ is the sample from the gaussian distribution $\mathcal{N}(z^t,\sigma^2)$. Moreover, $z^t$ is updated with stochastic gradient descent. The detailed procedure is presented in Algorithm \ref{alg:NES}. We do not need to do projection explicitly  since $\delta$ already satisfies $\|\delta\|_\infty < \varepsilon$. 

Next we shall briefly explain why applying NES on the embedding space $z$ can accelerate the search process. Adversarial examples can be viewed as a distribution lying around a given input. Usually this distribution is concentrated on some small regions, making the search process relatively slow. After training on the source network, the adversarial perturbations of {\GenAttack} would have high level semantic patterns that are likely to be adversarial patterns of the target network. Therefore searching over $z$ is like searching adversarial examples in a lower dimensional space containing likely adversarial patterns. The distribution of adversarial perturbations in this space is much less concentrated. It is thus much easier to find effective adversarial patterns in the embedding space.

\begin{algorithm}[h] 
\caption{Black-Box adversarial attack on the embedding space} 
\label{alg:NES} 
\begin{algorithmic}[1]
\REQUIRE ~~\\
Target Network $F_t$; Input $x$ and its label $y$ or the target class $t$; Encoder $\mathcal{E}$; Decoder $\mathcal{D}$; Standard deviation $\sigma$; Learning rate $\eta$; Sample size $b$; Iterations $T$; Bound for adversarial perturbation $\varepsilon$ \\
\ENSURE Adversarial perturbation $\delta$ \\ 
\STATE $z_0 = \mathcal{E}(x)$
\FOR{$ t = 1$ to $T$}
\STATE Sample Gaussian noise $\nu_1, \nu_2, \cdots, \nu_b \sim \mathcal{N}(z_{t-1}, \sigma^2)$
\STATE Calculate $\mathcal{L}_i = \mathcal{L}_\text{untarget}(x, y)$ or $ \mathcal{L}_\text{target}(x, t)$
\STATE Update $z_t = z_{t-1} - \frac{\eta}{b}\sum_{i=1}^b \mathcal{L}_i \nabla_{z_{t-1}} \log \mathcal{N}(\nu_i|z_{t-1}, \sigma^2)$
\ENDFOR
\RETURN $\delta = \varepsilon \tanh(\mathcal{D}(z_T))$
\end{algorithmic}
\end{algorithm}

\section{Experiments}
We evaluated the number of queries versus success rate of {\GenAttack} on undefended network in two datasets: MNIST \citep{mnist} and ImageNet \citep{imagenet}. Moreover, we evaluated the efficiency of our method on adversarially defended networks in CIFAR10 \citep{CIFAR10} and ImageNet. We also attacked Google Cloud Vision API to show {\GenAttack} can generalize to truly black-box model.\footnote{Our code is available at \url{https://github.com/TransEmbedBA/TREMBA}} We used the hinge loss from Eqn \ref{eqn:min} and \ref{eqn:min-target} as the surrogate loss for un-targeted and targeted attack respectively. 

We compared {\GenAttack} to four methods: (1) \textbf{NES}: Method introduced by \citep{nes}, but without the sign function for reasons explained earlier. (2) \textbf{Trans-NES}: Take an adversarial perturbation generated by PGD or FGSM on the source model to initialize NES. (3) \textbf{AutoZOOM}: Attack target network with an unsupervised autoencoder described in \citep{autozoom}. For fair comparisons with other methods, the strategy of choosing sample size was removed. (4) \textbf{P-RGF}: Prior-guided random gradient-free method proposed in \citep{prgf}. The $\text{P-RGF}_\text{D}(\lambda^*)$ version was compared. We also combined P-RGF with initialization from Trans-NES$_\text{PGD}$ to form a more efficient method for comparison, denoted by Trans-P-RGF.


Since different methods achieve different success rates, we need to compare their efficiency at different levels of success rate. For method $i$ with success rate $s_i$, the average number of queries is $q_i$ for all success examples. Let $q^*$ denote the upper limit of queries, we modified the average number of queries to be
$q_i^*=[(\max_{j} s_j - s_i)\cdot q^* + s_i \cdot q_i]/\max_{j} s_j$, which unified the level of success rate and treated queries of failure examples as the upper limit on the number of queries. Average queries sometimes could be misleading due to the the heavy tail distribution of queries. Therefore we plot the curve of success rate at different query levels to show the detailed behavior of different attacks. 

The upper limit on the number of queries was set to $50000$ for all datasets, which already gave very high success rate for nearly all the methods. Only correctly classified images were counted towards success rate and average queries. And to fairly compare these methods, we chose the sample size to be the same for all methods. We also added momentum and learning decay for optimization. And we counted the queries as one if its starting point successfully attacks the target classifier. The learning rate was fine-tuned for all algorithms. We listed the hyperparameters and architectures of generators and classifiers in Appendix \ref{sec:architecture} and \ref{sec:hyperparameter-list}.

\subsection{Black-box Attack on MNIST}
We trained four neural networks on MNIST, denoted by ConvNet1, ConvNet1*, ConvNet2 and FCNet. ConvNet1* and ConvNet1 have the same architecture but different parameters. All the network achieved about $99\%$ accuracy. The generator $\mathcal{G}$ was trained on ConvNet1* using all images from the training set. Each attack was tested on images from the MNIST test set. The limit of $\ell_{\infty}$ was $\varepsilon=0.2$.

\begin{figure}[t]
    \centering
    \includegraphics[width=0.8\linewidth]{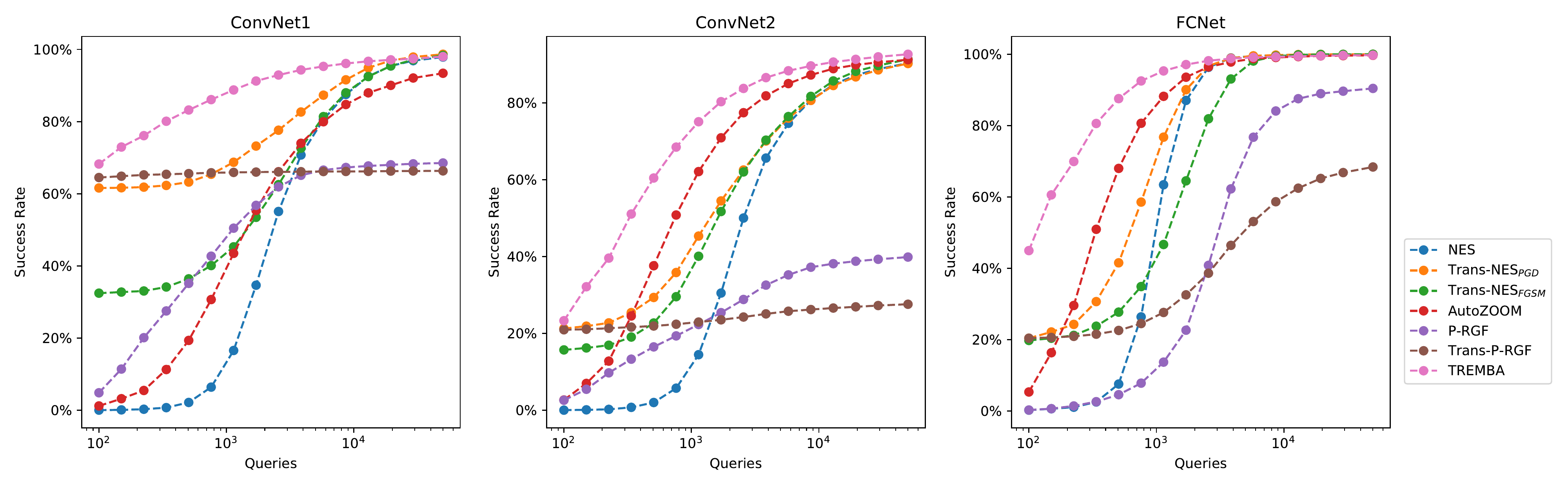}
    \label{fig:mnist}
    \caption{Success rate of un-targeted attack at different query levels for undefended MNIST models.}
\end{figure}

We performed un-targeted attack on MNIST. Table \ref{tab:mnist} lists the success rate and the average queries. Although the success rate of {\GenAttack} is slightly lower than Trans-NES in ConvNet1 and FCNet, their success rate are already close to $100\%$ and {\GenAttack} achieves about $50\%$ reduction of queries compared with other attacks. In contrast to efficient attack on ImageNet, P-RGF and Trans-P-RGF behaves very bad on MNIST. Figure \ref{fig:mnist} shows that {\GenAttack} consistently achieves higher success rate at nearly all query levels.

\begin{table}[t]
    \centering
    \caption{Success rate and average queries of un-targeted attack on MNIST. $\varepsilon=0.2$}
    \setlength\tabcolsep{10.0pt}
    \begin{tabular}{c c c c c c c}
        \hline
        \multirow{2}*{Attack} & 
        \multicolumn{2}{c}{ConvNet1} & \multicolumn{2}{c}{ConvNet2} & \multicolumn{2}{c}{FCNet} \\
        \cmidrule(r){2-3} \cmidrule(r){4-5} \cmidrule(r){6-7}
        & Success & Queries & Success  & Queries & Success & Queries \\
        \hline
        NES & 97.88\% & 4380 & 90.32\% & 5428 & 99.98\% & 1183 \\
        Trans-NES$_\text{PGD}$ & \textbf{98.65\%} & 2113 & 90.22\% & 4691 & \textbf{99.99\%} & 818 \\
        Trans-NES$_\text{FGSM}$ & 98.34\% & 3592 & 91.32\% & 4218 & 99.99\% & 1540   \\
        AutoZOOM & 93.39\% & 5874 & 91.21\% & 2645 & 99.69\% & 823 \\
        P-RGF & 68.53\% & 16135 & 39.85\% & 29692 & 90.42\% & 8289\\
        Trans-P-RGF & 66.34\% & 16428 & 27.57\% & 35576 & 68.39\% & 18818\\
        \hline
        {\GenAttack} & 98.00\% & \textbf{1064} & \textbf{92.63\%} & \textbf{1359} & 99.75\% & \textbf{470} \\
        \hline
    \end{tabular}
    \label{tab:mnist}
\end{table}

\begin{table}[t]
    \centering
    \setlength\tabcolsep{4.3pt}
    \caption{Success rate and average queries of black-box targeted attack on ImageNet. Targeted class is class 0 (tench). $\varepsilon=0.03125$}
    \begin{tabular}{c c c c c c c c c}
        \hline
        \multirow{2}*{Attack} & 
        \multicolumn{2}{c}{VGG19} & \multicolumn{2}{c}{Resnet34} & \multicolumn{2}{c}{DenseNet121} & \multicolumn{2}{c}{MobilenetV2} \\
        \cmidrule(r){2-3} \cmidrule(r){4-5} \cmidrule(r){6-7} \cmidrule(r){8-9}
        & Success & Queries & Success & Queries & Success & Queries & Success & Queries\\
        \hline
        NES & 94.86\% & 12283 & 93.89\% & 14418 & 95.65\% & 12538 & 97.76\% & 10276  \\
        Trans-NES$_\text{PGD}$ & 96.26\% & 6854 & 95.97\% & 8737 & 96.59\% & 8627 & 98.04\% & 9375 \\
        Trans-NES$_\text{FGSM}$ & 90.85\% & 12885 & 91.81\% & 14090 & 93.61\% & 12859 & 97.48\% & 9983 \\
        AutoZOOM & 25.80\% & 40195 & 26.25\% & 39681 & 31.98\% & 37628 & 27.03\% & 39689  \\
        P-RGF & 96.12\% & 6951 & 90.28\% & 10221 & 91.84\% & 11563 & 88.94\% & 14596 \\
        Trans-P-RGF & 98.06\% & 2262 & 93.61\% & 6309 & 94.69\% & 7263 & 91.60\% & 10048 \\
        \hline
        {\GenAttack} & \textbf{98.47\%} & \textbf{853} & \textbf{96.38\%} & \textbf{1206} & \textbf{98.50}\% & \textbf{1124} & \textbf{99.16\%} & \textbf{1210} \\
        \hline
    \end{tabular}
    \label{tab:imagenet-targeted}
\end{table}

\begin{figure}[t]
    \centering
    \centering
    \includegraphics[width=\linewidth]{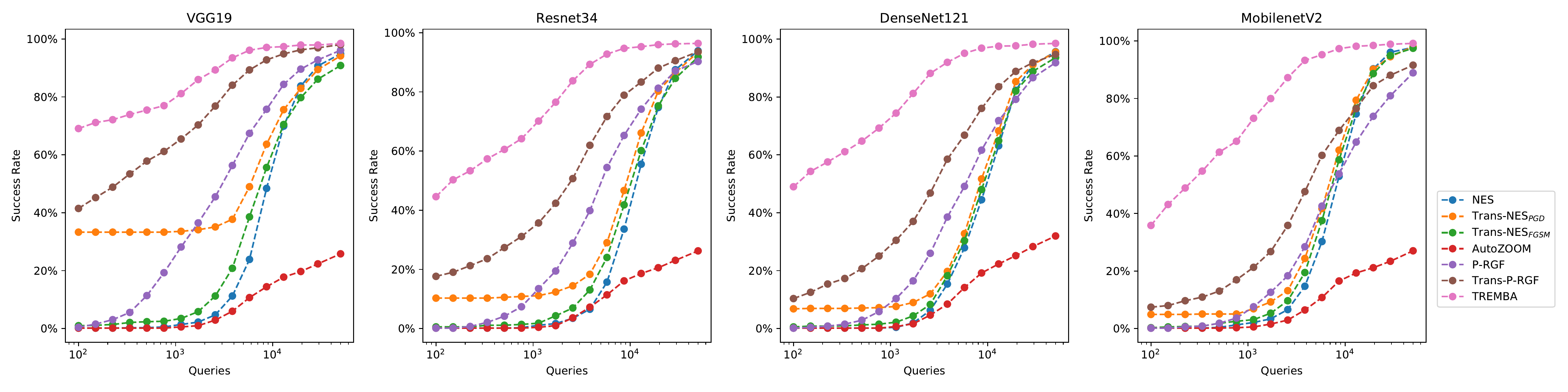}
    \caption{The success rate of black-box adversarial targeted  attack at different query levels for ImageNet models. The targeted class is tench}
    \label{fig:imagenet-target}

\end{figure}

\begin{figure}[!t]
    \centering
    \centering
    \includegraphics[width=\linewidth]{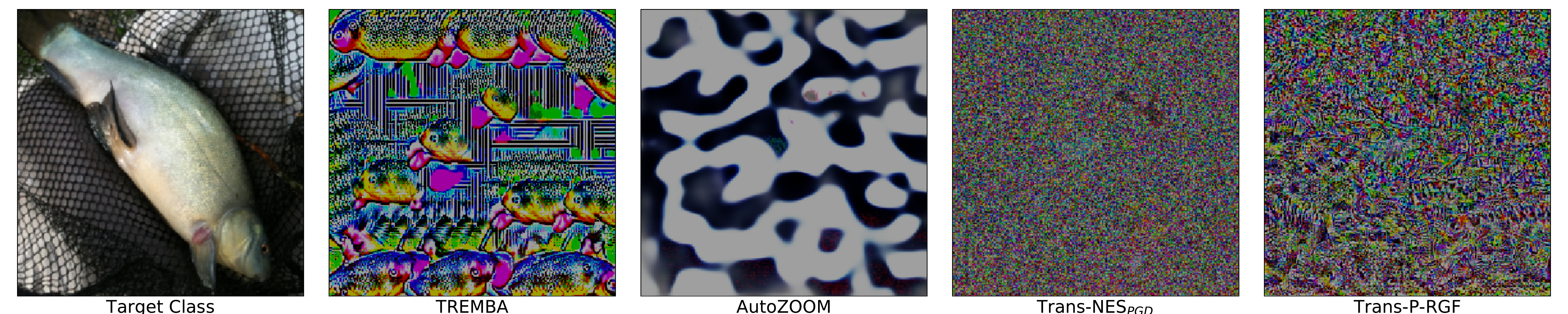}
    \caption{Visualization of adversarial perturbations targeted at tench}
    \label{fig:imagenet-visual-all}
\end{figure}

\subsection{Black-Box Attack on ImageNet}

We randomly divided the ImageNet validation set into two parts, containing $49000$ and $1000$ images respectively. The first part was used as the training data for the generator $\mathcal{G}$, and the second part was used for evaluating the attacks. We evaluated the efficiency of all adversarial attacks on VGG19 \citep{vgg}, Resnet34 \citep{resnet}, DenseNet121 \citep{densenet} and MobilenetV2 \citep{sandler2018mobilenetv2}. All networks were downloaded using \textit{torchvision} package. We set $\varepsilon=0.03125$.

Following \citep{ensembleattack}, we used an ensemble (VGG16, Resnet18, Squeezenet \citep{iandola2016squeezenet} and Googlenet \citep{googlenet}) as the source model to improve transferablity \citep{ensembleattack} for both targeted and un-targeted attack. {\GenAttack}, Trans-NES, P-RGF and Trans-P-RGF all used the same source model for fair comparison. 
We chose several target class. Here, we show the result of attacking class 0 (tench) in Table \ref{tab:imagenet-targeted} and Figure \ref{fig:imagenet-target}. And we leave the result of attacking other classes in Appendix \ref{sec:imageget-target}. The average queries for {\GenAttack} is about 1000 while nearly all the average queries for other methods are more than 6000. {\GenAttack} also achieves much lower queries for un-targeted attack on ImageNet. The result is shown in Appendix \ref{sec:imagenet-untarget} due to space limitation. And we also compared {\GenAttack} with CombOpt \citep{moon2019parsimonious} in the Appendix \ref{sec:combopt}.

Figure \ref{fig:imagenet-visual-all} shows the adversarial perturbations of different methods. Unlike adversarial perturbations produced by PGD, the perturbations of {\GenAttack} reflect some high level semantic patterns of the targeted class such as the fish scale. As neural networks usually capture such patterns for classification, the adversarial perturbation of {\GenAttack} would be more easy to transfer than the noise-like perturbation produced by PGD. Therefore {\GenAttack} can search very effectively for the target network. More examples of perturbations of {\GenAttack} are shown in Appendix \ref{sec:visual}.

\textbf{Choice of ensemble:} We performed attack on different ensembles of source model, which is shown in Appendix \ref{sec:ensemble}. {\GenAttack} outperforms the other methods in different ensemble model. And more source networks lead to better transferability for {\GenAttack}, Trans-NES and Trans-P-RGF. 


\textbf{Varying $\varepsilon$:} We also changed $\varepsilon$ and performed attack on $\varepsilon=0.02$ and $\varepsilon=0.04$. As shown in Appendix \ref{sec:epsilon}, {\GenAttack} still outperforms the other methods despite using the $\mathcal{G}$ trained on $\varepsilon=0.03125$. We also show the result of {\GenAttack} for commonly used $\varepsilon=0.05$.

\textbf{Sample size and dimension the embedding space:} To justify the choice of sample size, we performed a hyperparameter sweep over $b$ and the result is shown in Appendix \ref{sec:sample-size}. And we also changed the dimension of the embedding space for AutoZOOM and Trans-P-RGF. As shown in Appendix \ref{sec:dimz}, the performance gain of {\GenAttack} does not purely come from the diminishing of dimension of the embedding space.

\subsection{Black-Box Attack on Defended Models}

This section presents the results for attacking defended networks. We performed un-targeted attack on two SOTA defense methods on CIFAR10 and ImageNet. MNIST is not studied since it is already robust against very strong white-box attacks. For CIFAR10, the defense model was going through PGD minimax training \citep{madry}. We directly used their model as the source network\footnote{\url{https://github.com/MadryLab/cifar10_challenge}\label{footnode:cifar10_chanllenge}}, denoted by WResnet. To test whether these methods can transfer to a defended network with a different architecture, we trained a defended ResNeXt \citep{resnext} using the same method. For ImageNet, we used the SOTA model\footnote{\url{https://github.com/facebookresearch/ImageNet-Adversarial-Training}} from \citep{denosing}. We used "ResNet152 Denoise" as the source model and transfered adversarial perturbations to the most robust "ResNeXt101 DenoiseAll". Following the previous settings, we set $\varepsilon=0.03125$ for both CIFAR10 and ImageNet. 

\begin{figure}[t]
    \centering
    \subfigtopskip = 0pt
    \subfigure[CIFAR10]{
        \begin{minipage}[t]{0.36\linewidth}
            \centering
            \includegraphics[width=\linewidth]{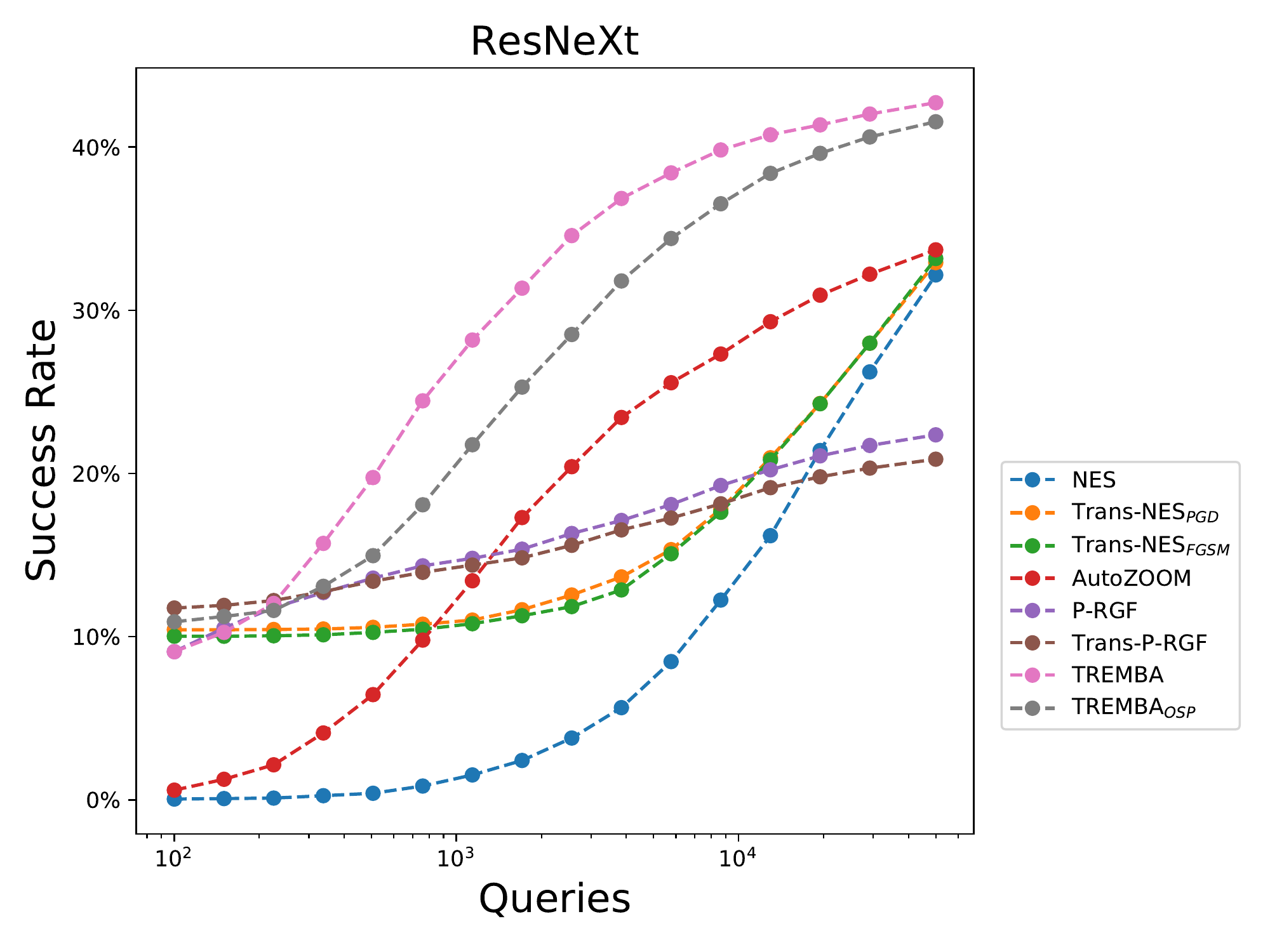}
        \end{minipage}
        \label{fig:cifar10_defense}
    }
    \subfigure[ImageNet]{
        \begin{minipage}[t]{0.36\linewidth}
            \centering
            \includegraphics[width=\linewidth]{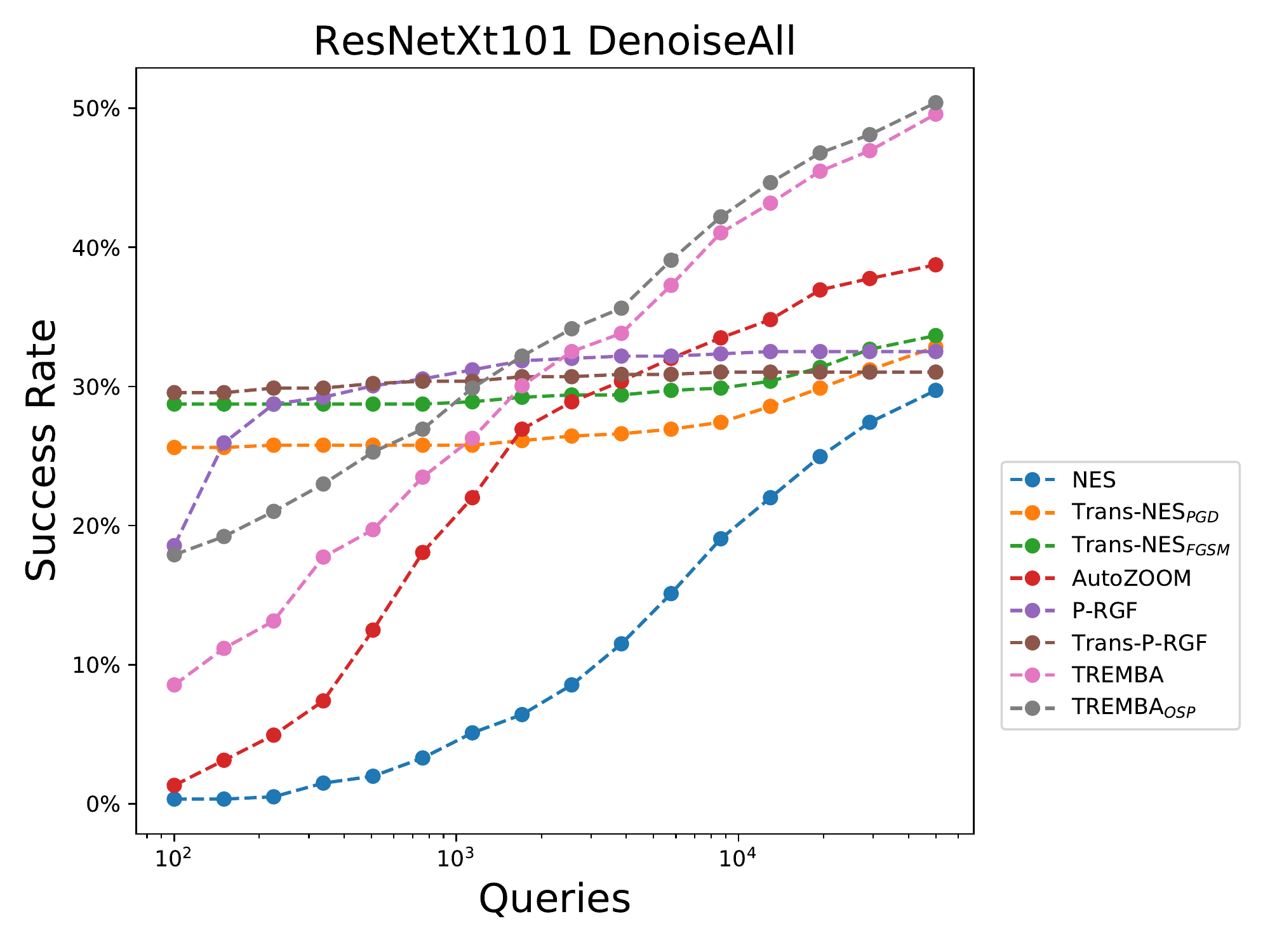}
        \end{minipage}
        \label{fig:imagenet_defense}
    }
    \caption{The success rate at different query levels for defended CIFAR10 and ImageNet models. (a)CIFAR10; (b)ImageNet.}
    \label{fig:defense}
\end{figure}

As shown in Table \ref{tab:defense}, {\GenAttack} achieves higher success rates with lower number of queries. {\GenAttack} achieves about $10\%$ improvement of success rate while the average queries are reduced by more than $50\%$ on ImageNet and by $80\%$ on CIFAR10. The curves in Figure \ref{fig:cifar10_defense} and \ref{fig:imagenet_defense} show detailed behaviors. The performance of AutoZOOM surpasses Trans-NES on defended models. We suspect that low-frequency adversarial perturbations produced by AutoZOOM will be more suitable to fool the defended models than the regular networks. However, the patterns learned by AutoZOOM are still worse than adversarial patterns learned by {\GenAttack} from the source network.

\textbf{An optimized starting point for {\GenAttack}}: $z_0 = \mathcal{E}(x)$ is already a good starting point for attacking undefended networks. However, the capability of generator is limited for defended networks \citep{wang2018a}. Therefore, $z_0$ may not be the best starting point we can get from the defended source network. To enhance the usefulness of the starting point, we optimized $z$ on the source network by gradient descent and found
\begin{equation}
    z_0^* = \mathop{\arg\min}_{z} \max \left(F_s(\varepsilon \tanh(\mathcal{D}(z))+x)_{y}-\max _{j \neq y_i} F_s(\varepsilon \tanh(\mathcal{D}(z))+x)_{j},-\kappa\right).
\end{equation}
The method is denoted by {\TGenAttack} ({\GenAttack} with optimized starting point). Figure \ref{fig:defense} shows {\TGenAttack} has higher success rate at small query levels, which means its starting point is better than {\GenAttack}.

\subsection{Attack Google Cloud Vision API}

We also attacked the Google Cloud Vision API, which was much harder to attack than the single neural network. Therefore we set $\varepsilon=0.05$ and perform un-targeted attack on the API, changing the top1 label to whatever is not on top1 before. We chose 10 images for the ImageNet dataset and set query limit to be 500 due to high cost to use the API. As shown Table \ref{tab:google-untarget}, {\GenAttack} achieves much higher accuracy success rate and lower number of queries. We show the example of successfully attacked image in Appendix \ref{sec:google-visual}.

\begin{table}[t]
    \centering
    \setlength\tabcolsep{4.0pt}
    \caption{Success rate of average queries of black-box un-targeted attack on defended CIFAR10 and ImageNet model. Source network is WResNet and ResNet152 Denoise.}
    \begin{tabularx}{\textwidth}{Y Y Y Y Y}
        \hline
        \multirow{2}*{Attack} & 
        \multicolumn{2}{c}{CIFAR10 ResneXt} & \multicolumn{2}{c}{ImageNet RexneXt101 DenoiseAll} \\
        \cmidrule(r){2-3} \cmidrule(r){4-5}
        & Success & Queries & Success & Queries  \\ \hline
        NES & 32.17\% & 24521 & 29.72\% & 26526\\
        Trans-NES$_\text{PGD}$ & 32.92\% & 20735& 32.84\% & 20446\\
        Trans-NES$_\text{FGSM}$ & 33.17\% & 20873 & 33.66\% & 18547 \\
        AutoZOOM & 33.70\% & 14870 & 38.75\% & 14605\\
        P-RGF & 22.37\% & 25818 & 32.51\% & 17926 \\
        Trans-P-RGF & 20.88\% & 27222 & 31.03\% & 19262\\
        \hline
        {\GenAttack} & \textbf{42.73\%} & \textbf{2528} & 49.59\% & 5985 \\
        {\TGenAttack} & 41.56\% & 4994 & \textbf{50.41\%} & \textbf{4771} \\
        \hline
    \end{tabularx}
    \label{tab:defense}
\end{table}

\begin{table}[!t]
    \centering
    \setlength\tabcolsep{7.0pt}
    \caption{Success rate and average queries of un-targeted attack of 10 images on Google Vision API.}
    \begin{tabular}{c c c c c c c}
        \hline
        Method &NES & AutoZOOM & Trans-NES$_\text{PGD}$ & P-RGF & Trans-P-RGF & {\GenAttack} \\ \hline
        Success & 70.00\% & 20.00\% & 70.00\% & 50.00\% & 60.00\% & 90.00\% \\
        Queries & 245 & 410 & 114 & 324 & 167 & 8\\
        \hline
    \end{tabular}
    \label{tab:google-untarget}
\end{table}


\section{Conclusion}

We propose a novel method, {\GenAttack}, to generate likely adversarial patterns for an unknown network. The method contains two stages: (1) training an encoder-decoder to generate adversarial perturbations for the source network; (2) search adversarial perturbations on the low-dimensional embedding space of the generator for any unknown target network. Compared with SOTA methods, {\GenAttack} learns an embedding space that is more transferable across different network architectures. It achieves two to six times improvements in black-box adversarial attacks on MNIST and ImageNet and it is especially efficient in performing targeted attack. Furthermore, {\GenAttack} demonstrates great capability in attacking defended networks, resulting in a nearly $10\%$ improvement on the attack success rate, with two to six times of reductions in the number of queries. {\GenAttack} opens up new ways to combine transfer-based and score-based attack methods to achieve higher efficiency in searching adversarial examples.

For targeted attack, TREMBA requires different generators to attack different classes. We believe methods from conditional image generation \citep{cgan} may be combined with TREMBA to form a single generator that could attack multiple targeted classes. We leave it as a future work.

\bibliography{ref.bib}
\bibliographystyle{iclr2020_conference}
\newpage
\appendix
\section{Experiment Result}
\subsection{Targeted Attack On ImageNet}
\label{sec:imageget-target}
\begin{figure}[!h]
    \centering
    \subfigure[Dipper]{
        \begin{minipage}[t]{\linewidth}
            \centering
            \includegraphics[width=0.94\linewidth]{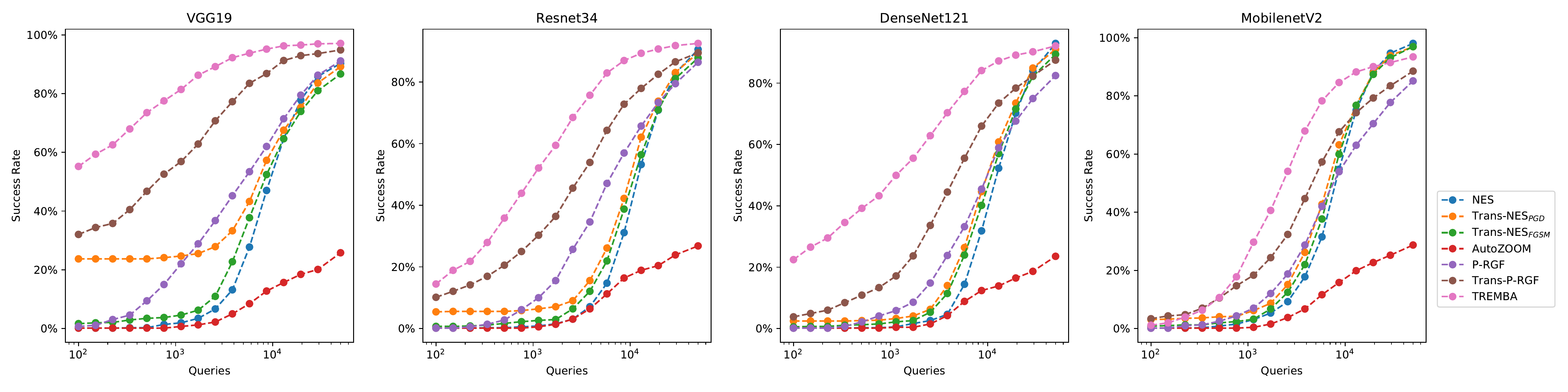}
        \end{minipage}
    }
    \subfigure[American chameleon]{
        \begin{minipage}[t]{\linewidth}
            \centering
            \includegraphics[width=0.94\linewidth]{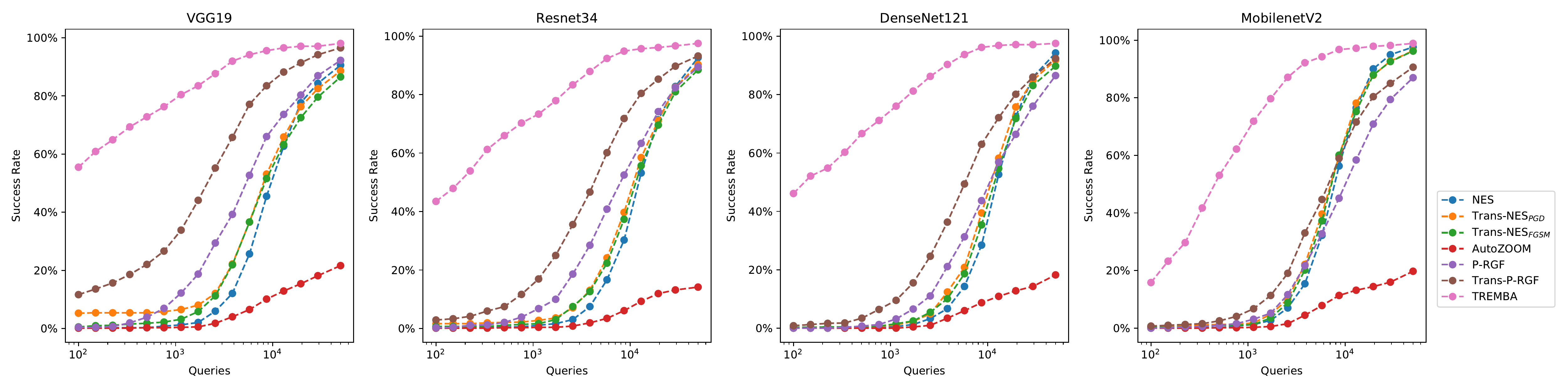}
        \end{minipage}
    }
    \subfigure[Night snake]{
        \begin{minipage}[t]{\linewidth}
            \centering
            \includegraphics[width=0.94\linewidth]{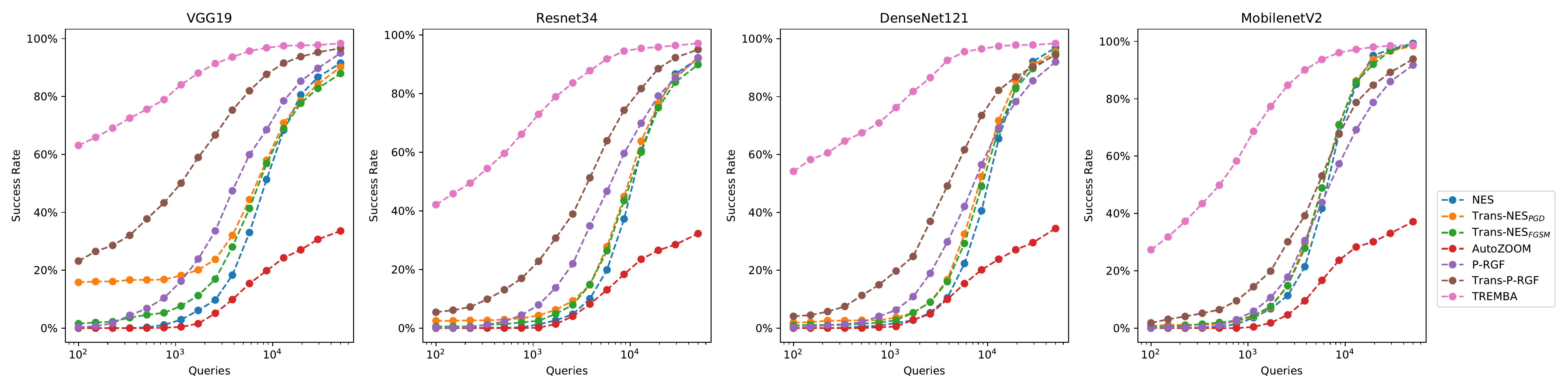}
        \end{minipage}
    }
    \subfigure[Ruffed grouse]{
        \begin{minipage}[t]{\linewidth}
            \centering
            \includegraphics[width=0.94\linewidth]{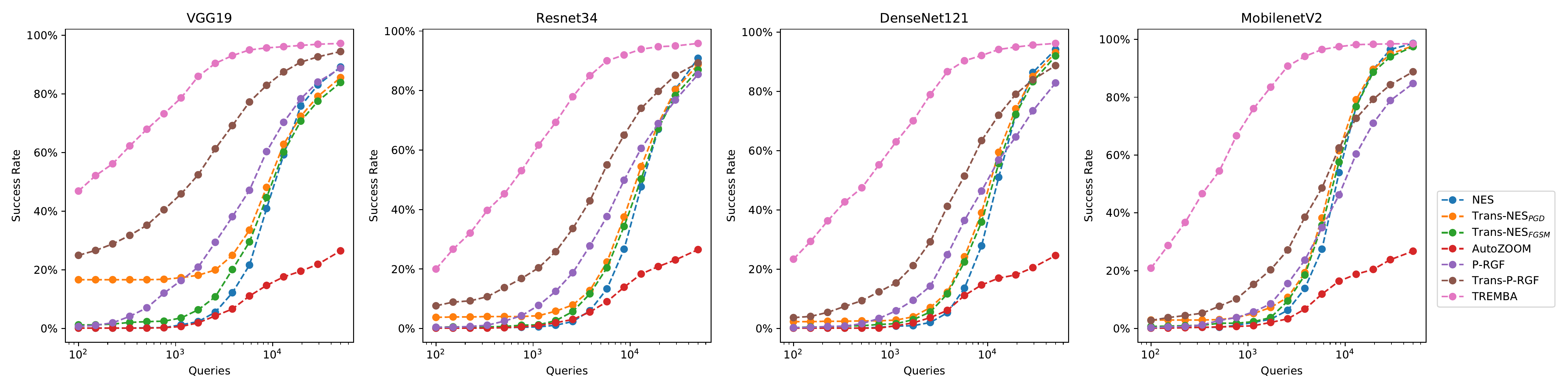}
        \end{minipage}
    }
    \subfigure[Black swan]{
        \begin{minipage}[t]{\linewidth}
            \centering
            \includegraphics[width=0.94\linewidth]{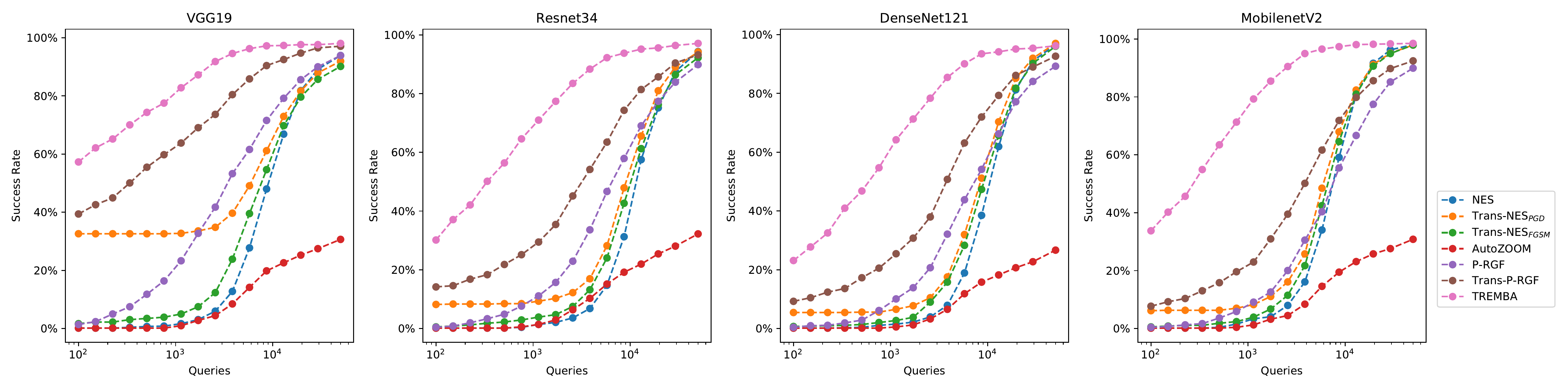}
        \end{minipage}
    }
    \caption{The success rate at different query levels for attack targeted at different class. Targeted classes are: (a)Dipper; (b)American chameleon; (c)Night snake; (d)Ruffed grouse; (e)Black swan}
    \label{fig:imagenet-target-appendix}
\end{figure}

Figure \ref{fig:imagenet-target-appendix} shows result of the targeted attack on dipper, American chameleon, night snake, ruffed grouse and black swan. {\GenAttack} achieves much higher success rate than other methods at almost all queries level.

\subsection{Un-targeted Attack on ImageNet}
\label{sec:imagenet-untarget}
\begin{figure}[!h]
    \centering
    \includegraphics[width=\linewidth]{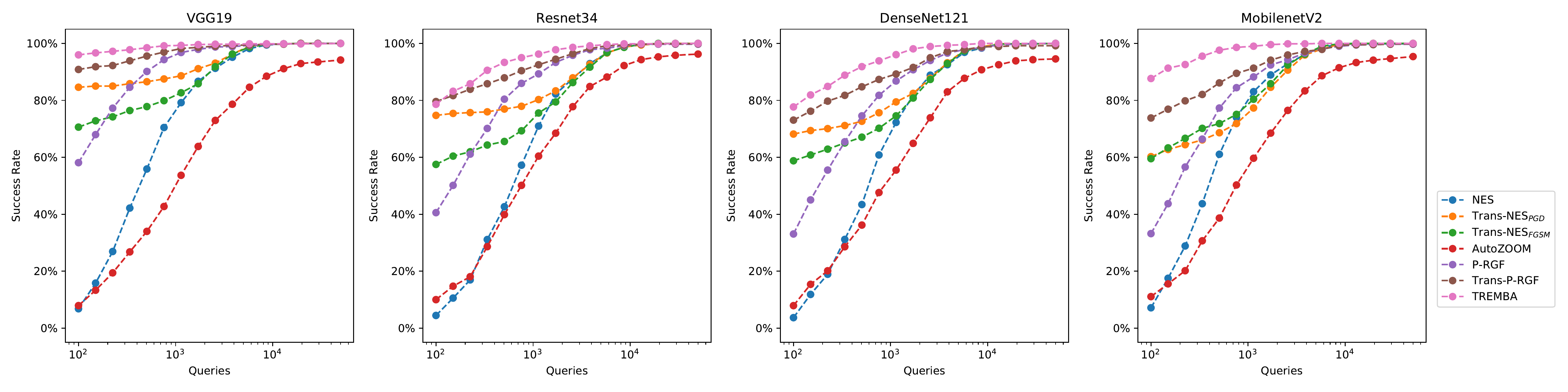}
    \caption{The success rate of un-targeted black-box adversarial attack at different query levels for undefended ImageNet models. }
    \label{fig:imagenet}
\end{figure}

We used the same source model from targeted attack as the source model for un-targeted attack. We report our evaluation results in Table \ref{tab:imagenet} and Figure \ref{fig:imagenet}. Compared with Trans-P-RGF, {\GenAttack} reduces the number of queries by more than a half in ResNet34, DenseNet121 and MobilenetV2. Searching in the embedding space of generator remains very effective even when the target network architecture differs significantly from the networks in the source model.

\begin{table}[!h]
    \centering
    \caption{Success rate and average queries of un-targeted attack on ImageNet. $\varepsilon=0.03125$}
    \setlength\tabcolsep{4.3pt}
    \begin{tabular}{c c c c c c c c c}
        \hline
        \multirow{2}*{Attack} & 
        \multicolumn{2}{c}{VGG19} & \multicolumn{2}{c}{Resnet34} & \multicolumn{2}{c}{DenseNet121} & \multicolumn{2}{c}{MobilenetV2} \\
        \cmidrule(r){2-3} \cmidrule(r){4-5} \cmidrule(r){6-7} \cmidrule(r){8-9}
        & Success & Queries & Success & Queries & Success & Queries & Success & Queries\\ \hline
        NES & \textbf{100\%} & 924 & \textbf{100\%} & 1255 & \textbf{100\%} & 1235 & 99.86\% & 872 \\
        Trans-NES$_\text{PGD}$ & \textbf{100\%} & 441 & \textbf{100\%} & 827 & \textbf{100\%} & 838 & \textbf{100\%} & 733  \\
        Trans-NES$_\text{FGSM}$ & \textbf{100\%} & 586 & \textbf{100\%} & 982 & \textbf{100\%} & 961 & \textbf{100\%} & 648 \\
        AutoZOOM & 94.18\% & 5184 & 96.25\% & 3754 & 94.56\% & 4567 & 95.38\% & 4213  \\
        P-RGF & \textbf{100\%} & 277 & 99.72\% & 635 & \textbf{100}\% & 709 & 99.72\% & 730 \\
        Trans-P-RGF & \textbf{100\%} & 130 & 99.86\% & 371 &  99.18\% & 806 & 99.86\% & 522 \\ \hline
        {\GenAttack} & \textbf{100\%} & \textbf{88} & \textbf{100\%} & \textbf{183} & \textbf{100\%} & \textbf{172} & \textbf{100\%} & \textbf{61} \\
        \hline
    \end{tabular}
    \label{tab:imagenet}
\end{table}

\subsection{Visualization of targeted perturbation}
\label{sec:visual}

\begin{figure}[!h]
    \centering
    \includegraphics[width=0.95\linewidth]{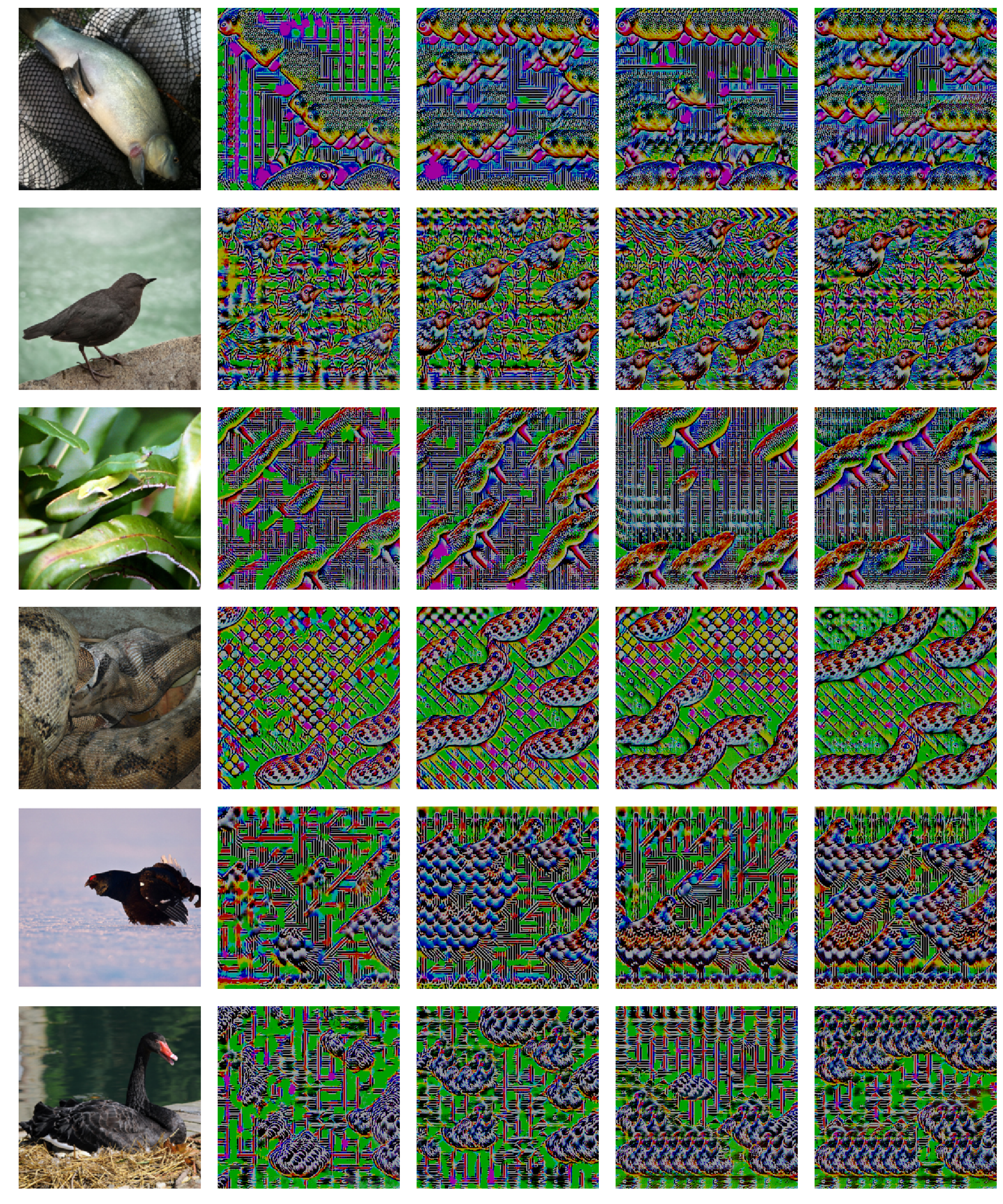}
    \caption{Visualization of adversarial perturbations for targeted attack on ImageNet. The first column shows one example of the target class. Other columns show the adversarial perturbations.}
    \label{fig:imagenet-visualize}
\end{figure}

Figure \ref{fig:imagenet-visualize} shows some examples of adversarial perturbations produced by {\GenAttack}. The first column is one image of the target class and other columns are examples of perturbations (amplified by 10 times). It is easy to discover some features of the target class in the adversarial perturbation such as the feather for birds and the body for snakes.

\subsection{Experiments on Different Ensembles}
\label{sec:ensemble}

\begin{figure}[!h]
    \centering
    \includegraphics[width=\linewidth]{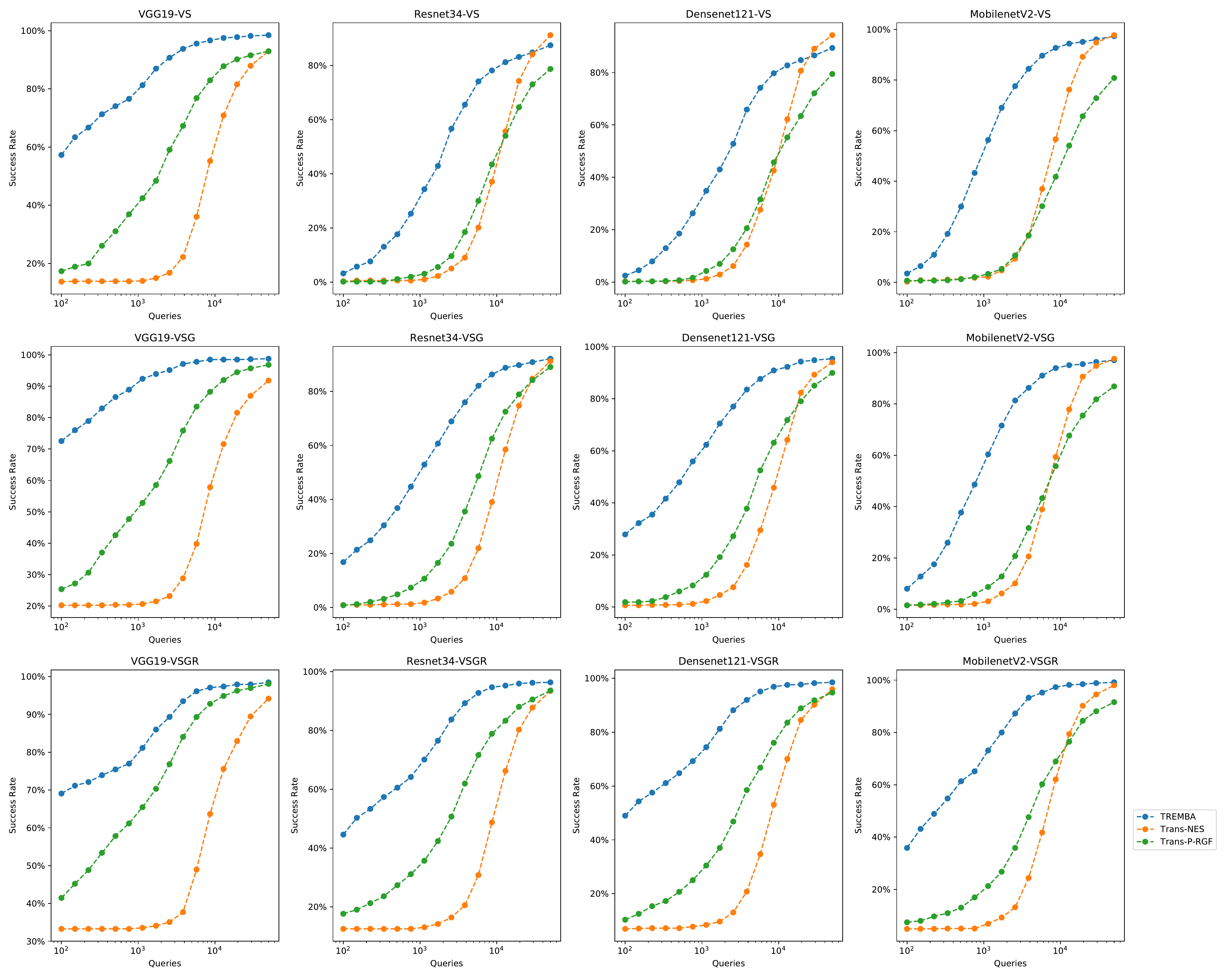}
    \caption{We show the success rate at different query levels for targeted attack for different ensemble source networks. V represents VGG16; S represents Squeezenet; G represents Googlenet; R represents Resnet18}
    \label{fig:imagenet-target-ensemble}
\end{figure}

We chose two more source ensemble models for evaluation. The first ensemble contains VGG16 and Squeezenet. And the second ensemble is consist of VGG16, Squeezenet and Googlenet. Figure \ref{fig:imagenet-target-ensemble} shows our result for targeted attack for ImageNet. We only compared Trans-NES$_\text{PGD}$ and Trans-P-RGF since they are the best variants from Trans-NES and P-RGF.

\subsection{Varying $\varepsilon$}
\label{sec:epsilon}
We chose $\varepsilon=0.02$ and $\varepsilon=0.04$ and performed targeted attack on ImageNet. Although {\GenAttack} used the same model that is trained on $\varepsilon=0.03125$, it still outperformed other methods, which shows that {\GenAttack} can also generalize to different strength of adversarial attack with different $\varepsilon$.

\begin{figure}[!h]
    \centering
    \subfigure[Targeted Attack $\varepsilon=0.02$]{
        \begin{minipage}[t]{0.9\linewidth}
            \centering
            \includegraphics[width=\linewidth]{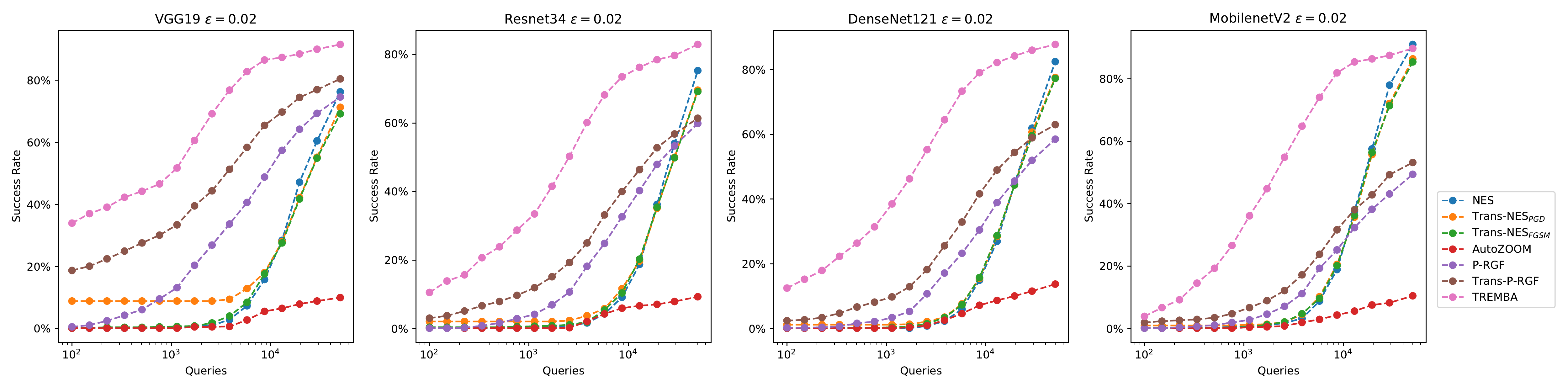}
        \end{minipage}
    }
    \subfigure[Targeted Attack $\varepsilon=0.04$]{
        \begin{minipage}[t]{0.9\linewidth}
            \centering
            \includegraphics[width=\linewidth]{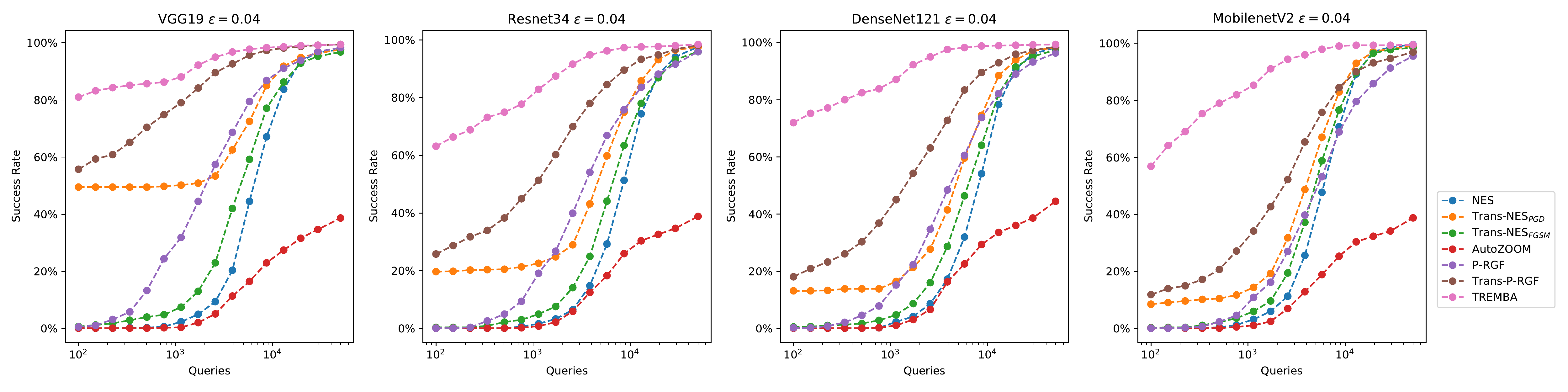}
        \end{minipage}
    }
    \caption{We show the success rate at different query levels for attack at different $\varepsilon$ for ImageNet.}
    \label{fig:imagenet-target-epsilon}
\end{figure}

\begin{figure}[!h]
    \centering
    \subfigure[Un-targeted Attack $\varepsilon=0.05$]{
        \begin{minipage}[t]{0.9\linewidth}
            \centering
            \includegraphics[width=\linewidth]{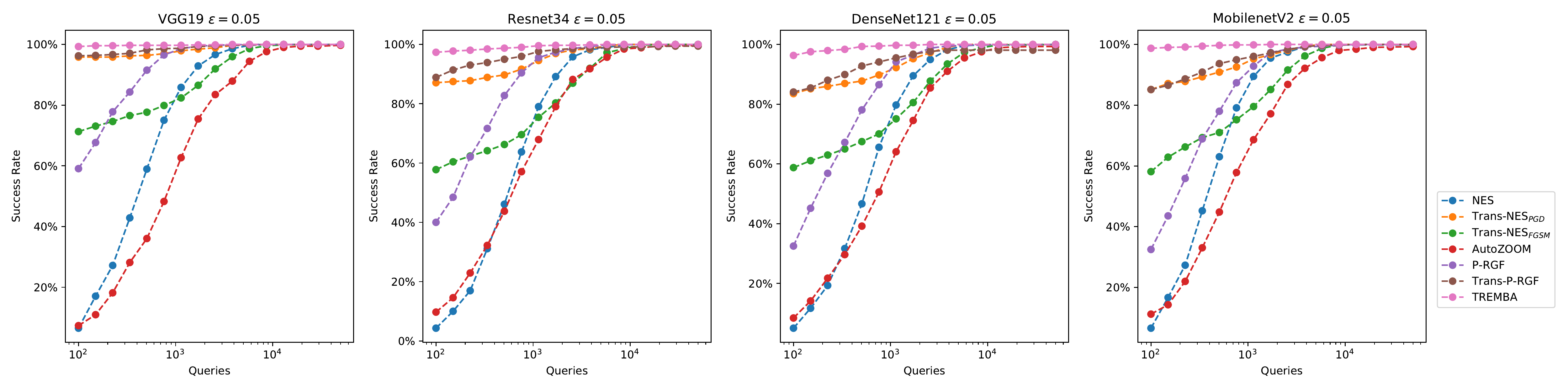}
        \end{minipage}
    }
    \subfigure[Targeted Attack $\varepsilon=0.05$]{
        \begin{minipage}[t]{0.9\linewidth}
            \centering
            \includegraphics[width=\linewidth]{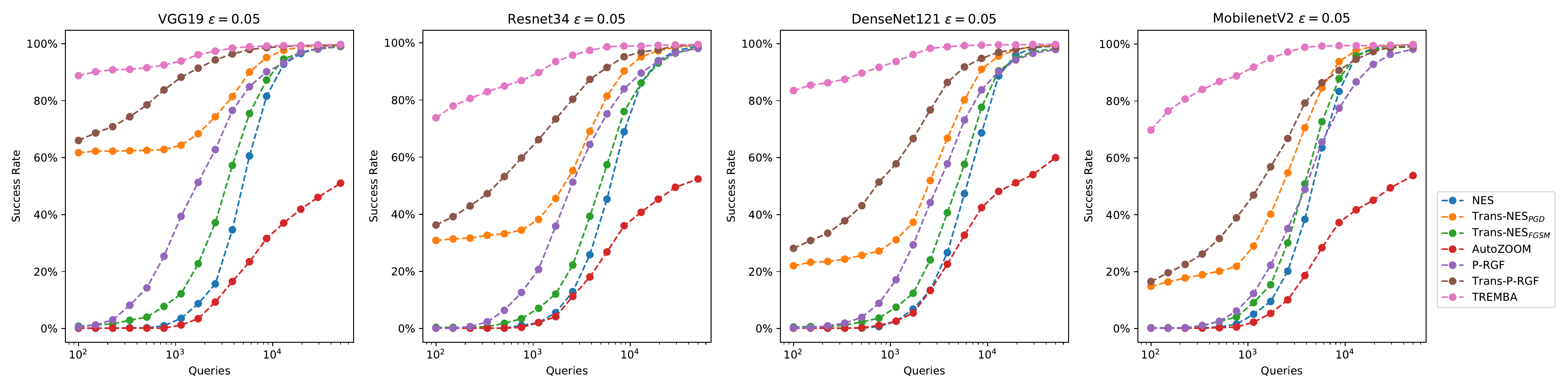}
        \end{minipage}
    }
    \caption{We show the success rate at different query levels for targeted and un-targeted attack at $\varepsilon=0.05$ for ImageNet.}
    \label{fig:imagenet-epsilon-005}
\end{figure}

For the commonly used $\varepsilon=0.05$, {\GenAttack} also performs well. The results are shown in Table \ref{tab:imagenet-untarget-005}, Table \ref{tab:imagenet-target-005}, and Figure \ref{fig:imagenet-epsilon-005}. 

\begin{table}[!h]
    \centering
    \caption{Success rate and average queries of un-targeted attack on ImageNet. $\varepsilon=0.05$}
    \setlength\tabcolsep{4.3pt}
    \begin{tabular}{c c c c c c c c c}
        \hline
        \multirow{2}*{Attack} & 
        \multicolumn{2}{c}{VGG19} & \multicolumn{2}{c}{Resnet34} & \multicolumn{2}{c}{DenseNet121} & \multicolumn{2}{c}{MobilenetV2} \\
        \cmidrule(r){2-3} \cmidrule(r){4-5} \cmidrule(r){6-7} \cmidrule(r){8-9}
        & Success & Queries & Success & Queries & Success & Queries & Success & Queries\\ \hline
        NES & \textbf{100\%} & 651& \textbf{100\%} & 850 & \textbf{100\%} & 840 & 99.86\% & 640 \\
        Trans-NES$_\text{PGD}$ & \textbf{100\%} & 74 & \textbf{100\%} & 196 & \textbf{100\%} & 235 & \textbf{100\%} & 169  \\
        Trans-NES$_\text{FGSM}$ & \textbf{100\%} & 232 & \textbf{100\%} & 401 & \textbf{100\%} & 361 & \textbf{100\%} & 272 \\
        AutoZOOM & 99.72\% & 1743 & 99.58\% & 1481 & 99.32\% & 1730 & 99.29\% & 1672  \\
        P-RGF & \textbf{100\%} & 178 & \textbf{100\%} & 328 & \textbf{100\%} & 436 & \textbf{100\%} & 402 \\
        Trans-P-RGF & \textbf{100\%} & 44 & 99.44\% & 418 &  98.09\% & 1049 & \textbf{100\%} & 157 \\ \hline
        {\GenAttack} & \textbf{100\%} & \textbf{8} & \textbf{100\%} & \textbf{27} & \textbf{100\%} & \textbf{19} & \textbf{100\%} & \textbf{8} \\
        \hline
    \end{tabular}
    \label{tab:imagenet-untarget-005}
    \caption{Success rate and average queries of targeted attack on ImageNet. $\varepsilon=0.05$}
    \setlength\tabcolsep{4.3pt}
    \begin{tabular}{c c c c c c c c c}
        \hline
        \multirow{2}*{Attack} & 
        \multicolumn{2}{c}{VGG19} & \multicolumn{2}{c}{Resnet34} & \multicolumn{2}{c}{DenseNet121} & \multicolumn{2}{c}{MobilenetV2} \\
        \cmidrule(r){2-3} \cmidrule(r){4-5} \cmidrule(r){6-7} \cmidrule(r){8-9}
        & Success & Queries & Success & Queries & Success & Queries & Success & Queries\\ \hline
        NES & 99.03\% & 6364& 98.89\% & 8003 & 99.32\% & 7525 & 99.72\$ & 5610 \\
        Trans-NES$_\text{PGD}$ & 99.31\% & 1968 & 99.31\% & 3549 & 99.46\% & 3731 & 99.86\% & 3223  \\
        Trans-NES$_\text{FGSM}$ & 99.03\% & 4997 & 98.33\% & 7298 & 98.23\% & 6874 & 99.16\% & 5034  \\
        AutoZOOM & 51.04\% & 30032 & 52.36\% & 28547 & 60.00\% & 25836 & 53.78\% & 28356  \\
        P-RGF & 99.17\% & 3704 & 98.05\% & 5498 & 97.96\% & 5769 & 98.17\% & 6896  \\
        Trans-P-RGF & 99.58\% & 662 & 99.31\% & 1896 & 99.05\% & 2267& 99.16\% & 3192  \\ \hline
        {\GenAttack} & \textbf{99.72\%} & \textbf{285} & \textbf{99.44\%} & \textbf{443} & \textbf{99.72\%} & \textbf{224} & \textbf{99.72\%} & \textbf{422} \\
        \hline
    \end{tabular}
    \label{tab:imagenet-target-005}
\end{table}

\subsection{Varying sample size}
\label{sec:sample-size}
We performed a hyperparameter sweep over $b$ on Densenet121 on un-targeted attack on ImageNet. $b=20$ may not be the best choice Trans-NES, but it is not the best for {\GenAttack}, either. Generally, the performance is not very sensitive to $b$, and {\GenAttack} will also outperform other methods even if we fine-tune the sample size for all the methods.

\begin{table}[h!]
    \centering
    \setlength\tabcolsep{4.3pt}
    \caption{Hyperparameter sweep over $b$ on Densenet121 for un-targeted attack on ImageNet}
    \begin{tabular}{c c c c c c c c c c c}
        \hline
        \multirow{2}*{Sweep over $b$} & 
        \multicolumn{2}{c}{$b=10$} & \multicolumn{2}{c}{$b=30$} & \multicolumn{2}{c}{$b=40$} & \multicolumn{2}{c}{$b=50$} \\
        \cmidrule(r){2-3} \cmidrule(r){4-5} \cmidrule(r){6-7} \cmidrule(r){8-9}
        & Success & Queries & Success & Queries & Success & Queries & Success & Queries\\ \hline
        NES & {100\%} & 1323 & {100\%} & 1284 & {100\%} &1433 & {100\%} & 1639 \\
        Trans-NES$_\text{PGD}$  & {100\%} & 915 & {100\%} & 791 & {100\%} & 707 & {100\%} & 639 \\
        Trans-NES$_\text{FGSM}$ & {100\%} & 1037 & {100\%} & 916 & {100\%} & 879 & {100\%} & 886 \\
        AutoZOOM & 90.9\% & 6052 & 96.2\% & 4148 & 97.1\% & 4066 & 97.3\% & 4366 \\
        P-RGF & 99.73\% & 717 & 99.86\% & 860 & 99.86\% & 949 & 99.86\% & 1095 \\
        Trans-P-RGF & 98.50\% & 1139 & 99.86\% & 479 & 99.86\% & 487 & {100\%} & 427 \\
        {\GenAttack} & {100\%} & 150 & {100\%} & 205 & {100\%} & 274 & {100\%} & 299\\

        \hline
    \end{tabular}
\end{table}

\subsection{dimension of the embedding space}
\label{sec:dimz}

We slightly changed the architecture of the autoencoder by adding max pooling layers and changing the number of filters and perform un-targeted attack on ImageNet. More specifically, we added additional max pooling layers after the first and the fourth convolution layers and changed the number of filters of the last layer in the encoder to be 8. Thus, the dimension of the embedding space would be $8\times 8\times 8$. And we also changed the factor of bilinear sampling in the decoder. The remaining settings are the same in Appendix \ref{sec:imagenet-untarget}. As shown in Table \ref{tab:dimz}, this autoencoder is even worse than the original autoencoder despite small dimension of the embedding space. In addition, we also changed to dimension of the data-dependent prior of Trans-P-RGF to match the dimension of {\GenAttack}, whose performance is also not better than before. They show that simply diminishing the size of the embedding space may not lead to better performance. The performance gain of {\GenAttack} comes beyond the effect of diminishing the dimension of the embedding space.

\begin{table}[!t]
    \centering
    \caption{Change of dimension of the embedding space of AutoZOOM. The task is un-targeted attack on ImageNet.}
    \setlength\tabcolsep{5.0pt}
    \begin{tabular}{c c c c c c c c c}
        \hline
        \multirow{2}*{Attack} & 
        \multicolumn{2}{c}{VGG19} & \multicolumn{2}{c}{Resnet34} & \multicolumn{2}{c}{DenseNet121} & \multicolumn{2}{c}{MobilenetV2} \\
        \cmidrule(r){2-3} \cmidrule(r){4-5} \cmidrule(r){6-7} \cmidrule(r){8-9}
        & Success & Queries & Success & Queries & Success & Queries & Success & Queries\\ \hline
        AutoZOOM & 64.35\% & 19684 & 71.94\% & 16931 &68.44\% & 17871 & 71.15\% & 16134\\  
        Trans-P-RGF & 99.86\% & 194 & 99.58\% & 508 & 99.59\% & 610 & 99.58\% & 705 \\
        \hline
        \end{tabular}
    \label{tab:dimz}
\end{table}

\subsection{Examples of Attacking Google Cloud Vision API}
\label{sec:google-visual}
\begin{figure}[!h]
    \centering
    \subfigure[Origin Image]{
        \begin{minipage}[t]{0.3\linewidth}
            \centering
            \includegraphics[width=\linewidth]{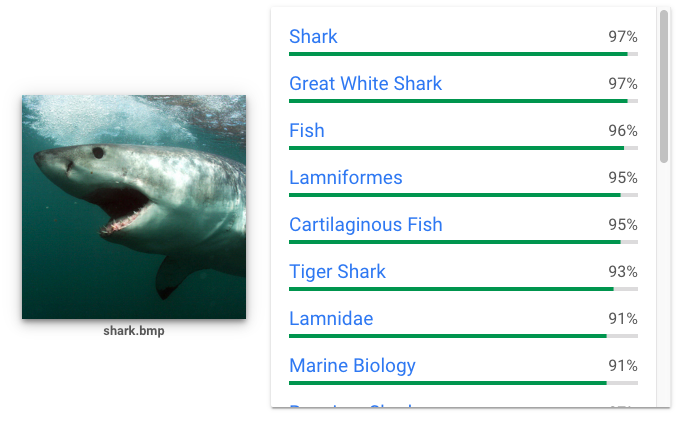}
        \end{minipage}
    }
    \subfigure[\GenAttack]{
        \begin{minipage}[t]{0.3\linewidth}
            \centering
            \includegraphics[width=\linewidth]{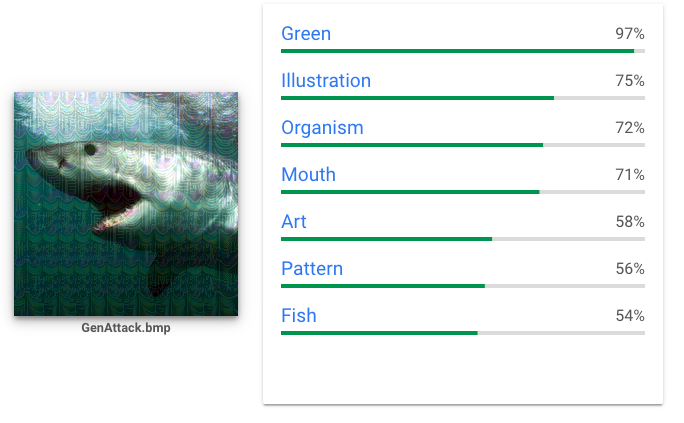}
        \end{minipage}
    }
    \subfigure[Trans-NES$_\text{PGD}$]{
        \begin{minipage}[t]{0.3\linewidth}
            \centering
            \includegraphics[width=\linewidth]{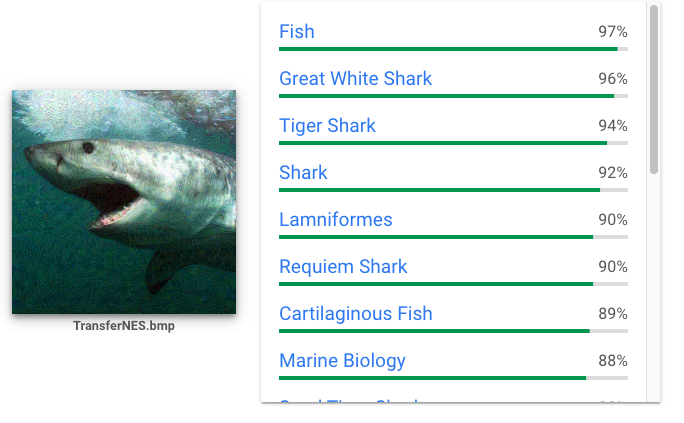}
        \end{minipage}
    }
    \caption{One example of adversarial image for attacking Google Cloud Vision API}
    \label{fig:google-visual}
\end{figure}

Figure \ref{fig:google-visual} shows one example of attacking Google Cloud Vision API. {\GenAttack} successfully make the shark to be classified as green. Compared with Trans-NES$_\text{PGD}$, {\GenAttack} hugely changes the labels of the image. It is hard to say the overall classification of Trans-NES$_\text{PGD}$ is wrong. However, the labels of {\GenAttack} are definitely not correct.

\newpage

\subsection{Comparision between {\GenAttack} and CombOpt}
\label{sec:combopt}

CombOpt is one of the SOTA score-based black-box attack. We compared our method with it on the targeted and un-targeted attack on Imagenet. The targeted attack is 0 and $\varepsilon=0.03125$. As shown in Table \ref{tab:combopt-untarget} and Table \ref{tab:combopt-target}, TREMBA requires much lower queries than CombOpt. It demonstrates the great improvement by combining the transfer-based and score-based attack.

\begin{table}[!t]
    \centering
    \caption{Comparision between CombOpt and {\GenAttack} for un-targeted attack on Imagenet.}
    \setlength\tabcolsep{5.0pt}
    \begin{tabular}{c c c c c c c c c}
        \hline
        \multirow{2}*{Attack} & 
        \multicolumn{2}{c}{VGG19} & \multicolumn{2}{c}{Resnet34} & \multicolumn{2}{c}{DenseNet121} & \multicolumn{2}{c}{MobilenetV2} \\
        \cmidrule(r){2-3} \cmidrule(r){4-5} \cmidrule(r){6-7} \cmidrule(r){8-9}
        & Success & Queries & Success & Queries & Success & Queries & Success & Queries\\ \hline
        CombOpt & \textbf{100\%} & 567 & \textbf{100\%} & 499 & \textbf{100\%} & 569 & \textbf{100\%} & 522\\  
        {\GenAttack} & \textbf{100\%} & \textbf{88} & \textbf{100\%} & \textbf{183} & \textbf{100\%} & \textbf{172} & \textbf{100\%} & \textbf{61} \\
        \hline
    \end{tabular}
    \label{tab:combopt-untarget}
    \caption{Comparision between CombOpt and {\GenAttack} for targeted attack on Imagenet.}
    \setlength\tabcolsep{5.0pt}
    \begin{tabular}{c c c c c c c c c}
        \hline
        \multirow{2}*{Attack} & 
        \multicolumn{2}{c}{VGG19} & \multicolumn{2}{c}{Resnet34} & \multicolumn{2}{c}{DenseNet121} & \multicolumn{2}{c}{MobilenetV2} \\
        \cmidrule(r){2-3} \cmidrule(r){4-5} \cmidrule(r){6-7} \cmidrule(r){8-9}
        & Success & Queries & Success & Queries & Success & Queries & Success & Queries\\ \hline
        CombOpt & 93.76\% & 9767 & 94.86\% & 8024 & 97.41\% & 6970 & 96.92\% & 8575\\  
        {\GenAttack} & \textbf{98.47\%} & \textbf{853} & \textbf{96.38\%} & \textbf{1206} & \textbf{98.50}\% & \textbf{1124} & \textbf{99.16\%} & \textbf{1210} \\
        \hline
    \end{tabular}

    \label{tab:combopt-target}
\end{table}

\section{Architecture of Classifiers and Generators}
\label{sec:architecture}
\subsection{Classifier}
\begin{table}[h]
    \centering
    \caption{Model architectures for the MNIST}
    \begin{tabular}{c c c} \hline
         ConvNet1 & ConvNet2 & FCNet  \\ \hline 
         Conv(64, 5, 5)+ReLU & Conv(16, 3, 3)+ReLU & FC(512)+ReLU \\
         MaxPool(2,2) & Conv(16, 3, 3)+ReLU & FC(10)+Softmax \\
         Conv(64, 5, 5)+ReLU & MaxPool(2,2) & \\
         MaxPool(2,2) & Conv(32, 3, 3)+ReLU & \\
         Dropout(0.25) & Conv(32, 3, 3)+ReLU & \\
         FC(128)+ReLU & Conv(32, 3, 3)+ReLU & \\
         Dropout(0.5) & MaxPool(2,2) & \\
         FC(10)+Softmax & FC(512)+ReLU & \\
         & FC(10)+Softmax & \\ \hline
    \end{tabular}
    \label{tab:mnist_model}
\end{table}

Table \ref{tab:mnist_model} lists the architectures of ConvNet1, ConvNet2 and FCNet. The architecture of ResNeXt used in CIFAR10 is from \url{https://github.com/prlz77/ResNeXt.pytorch}. We set the depth to be 20, the cardinality to be 8 and the widen factor to be 4. Other architectures of classifiers are specified in the corresponding paper.

\subsection{Generator}

\begin{table}[h]
    \centering
    \caption{Architectures of encoder and decoder. ConvReLUBN and DeconvReLUBN represent convolution or deconvolution followed by ReLU and batch normalization. The parameters ($c,m,n$) used in ConvReLUBN or DeconvReLUBN mean $c$ channels with $m\times n$ kernel size. MaxPool($m,n$) represents max pooling with ($m,n$) kernel size and ($m,n$) stride.}
    \begin{tabular}{c c c c}
        \hline
          & MNIST & CIFAR10 & ImageNet\\\hline
          \multirow{12}*{Encoder} & ConvReLUBN(16,3,3) & ConvReLUBN(16,3,3) & ConvReLUBN(16,3,3) \\
          & ConvReLUBN(32,3,3) & ConvReLUBN(32,3,3) & ConvReLUBN(32,3,3) \\
          & ConvReLUBN(32,3,3) & ConvReLUBN(32,3,3) & MaxPool(2,2) \\
          & MaxPool(2,2) &  MaxPool(2,2) &  ConvReLUBN(64,3,3)\\
          & ConvReLUBN(32,3,3) & ConvReLUBN(32,3,3) &  ConvReLUBN(64,3,3) \\
          & ConvReLUBN(16,3,3) &  ConvReLUBN(32,3,3) &  MaxPool(2,2)\\
          & ConvReLUBN(2,3,3) & ConvReLUBN(8,3,3) &   ConvReLUBN(128,3,3)\\
          & MaxPool(2,2) &  MaxPool(2,2) & ConvReLUBN(128,3,3)\\ 
          & & & MaxPool(2,2)\\
          & & & ConvReLUBN(32,3,3) \\
          & & & ConvReLUBN(8,3,3) \\
          & & & MaxPool(2,2) \\ \hline
          
         \multirow{9}*{Decoder} & ConvReLUBN(32,3,3) &  ConvReLUBN(32,3,3) & ConvReLUBN(32,3,3) \\
         & ConvReLUBN(32,3,3) &  ConvReLUBN(32,3,3) & DeconvReLUBN(64,3,3)\\
         & DeconvReLUBN(64,3,3) &  DeconvReLUBN(64,3,3) & ConvReLUBN(128,3,3)\\
         & ConvReLUBN(64,3,3) &  ConvReLUBN(64,3,3) & DeconvReLUBN(128,3,3)\\
         & ConvReLUBN(64,3,3) &  ConvReLUBN(64,3,3) & ConvReLUBN(128,3,3)\\
         & DeconvReLUBN(16,3,3) &  DeconvReLUBN(16,3,3) & DeconvReLUBN(64,3,3)\\
         & Conv(1,1,1) & Conv(3,1,1) & ConvReLUBN(32,3,3)\\
         & & & DeconvReLUBN(16,3,3) \\
         & & & ConvReLUBN(3,1,1) \\\hline
    \end{tabular}
    \label{tab:architecture}
\end{table}

Table \ref{tab:architecture} lists the architectures of generator for three datasets. For AutoZOOM, we find our architectures are not suitable and use the same generators in the corresponding paper.
\section{Hyperparameters}
\label{sec:hyperparameter-list}
\subsection{Training Generator}

We trained the generators with learning rate starting at 0.01 and decaying half every 50 epochs. The whole training process was 500 epochs. The batch size was determined by the memory of GPU. Specifically, we set batch size to be 256 for MNIST and CIFAR10 defense model, 64 for ImageNet model. All large $\kappa$ will work well for our method and we chose $\kappa=200.0$. All the experiments were performed using \textit{pytorch} on NVIDIA RTX 2080Ti.

\subsection{Evaluation}

Table \ref{tab:hyper_nes} to \ref{tab:hyper_embeddingos} list the hyperparameters for all the algorithms. The learning rate was fine-tuned for all the algorithms. We set sample size $b=20$ for all the algorithms for fair comparisons.

\begin{table}[h]
    \centering
    \setlength\tabcolsep{7.5pt}
    \caption{Hyperparameters for NES}
    \begin{tabular}{c c c c c c} \hline
        & \multirow{2}*{MNIST} & \multirow{2}*{CIFAR10} & \multicolumn{3}{c}{ImageNet} \\
        \cmidrule(r){4-6}
        & & & Un-targeted & Targeted & Un-targeted Defense \\ \hline
        Sample size ($b$) & 20 & 20 & 20 & 20 & 20 \\
        Learning rate ($\eta$)& 0.2 & 0.05 & 0.1 & 0.05 &  0.1 \\ \hline
    \end{tabular}

    \label{tab:hyper_nes}
\end{table}

\begin{table}[h]
    \centering
    \setlength\tabcolsep{5.5pt}
    \caption{Hyperparameters for Trans-NES$_\text{PGD}$ and Trans-NES$_\text{FGSM}$. White-box iteration, white-box margin and white-box learning rate mean the hyperparameters for generating the starting point on the source network for Trans-NES$_\text{PGD}$.}
    \begin{tabular}{c c c c c c} \hline
        & \multirow{2}*{MNIST} & \multirow{2}*{CIFAR10} & \multicolumn{3}{c}{ImageNet} \\
        \cmidrule(r){4-6}
        & & & Un-targeted & Targeted & Un-targeted Defense \\ \hline
        Sample size ($b$)& 20 & 20 & 20 & 20 & 20 \\
        Learning rate ($\eta$)& 0.2 & 0.05 & 0.1 & 0.05 & 0.1 \\ 
        White-box iteration & 50 & 100 & 50 & 50 & 100 \\
        White-box margin($\kappa$) & 100 & 100 & 100 & 100 & 100 \\
        White-box learning rate & 0.05 & 0.1 & 0.01 & 0.005 & 0.1 \\\hline
    \end{tabular}
    \label{tab:hyper_transnes}
\end{table}

\begin{table}[!h]
    \centering
    \setlength\tabcolsep{7.5pt}
    \caption{Hyperparameters for AutoZOOM.}
    \begin{tabular}{c c c c c c} \hline
        & \multirow{2}*{MNIST} & \multirow{2}*{CIFAR10} & \multicolumn{3}{c}{ImageNet} \\
        \cmidrule(r){4-6}
        & & & Un-targeted & Targeted & Un-targeted Defense \\ \hline
        Sample size ($b$)& 20 & 20 & 20 & 20 & 20 \\
        Learning rate ($\eta$)& 5.0 & 20.0 & 5.0 & 3.0 & 5.0 \\ \hline
    \end{tabular}
    \label{tab:hyper_autozoom}
\end{table}

\begin{table}[!h]
    \centering
    \setlength\tabcolsep{5.5pt}
    \caption{Hyperparameters for P-RGF and Trans-P-RGF.}
    \begin{tabular}{c c c c c c} \hline
        & \multirow{2}*{MNIST} & \multirow{2}*{CIFAR10} & \multicolumn{3}{c}{ImageNet} \\
        \cmidrule(r){4-6}
        & & & Un-targeted & Targeted & Un-targeted Defense \\ \hline
        Sample size ($b$) & 20 & 20 & 20 & 20 & 20 \\
        Learning rate ($\eta$)& 0.1 & 0.05 & 0.005 & 0.003 & 0.005 \\ 
        White-box iteration & 50 & 100 & 50 & 50 & 100 \\
        White-box margin($\kappa$) & 100 & 100 & 100 & 100 & 100 \\
        White-box learning rate & 0.05 & 0.1 & 0.01 & 0.01 & 0.1 \\\hline
    \end{tabular}

    \label{tab:hyper_pgf}
\end{table}

\begin{table}[!h]
    \centering
    \setlength\tabcolsep{7.5pt}
    \caption{Hyperparameters for {\GenAttack}.}
    \begin{tabular}{c c c c c c} \hline
        & \multirow{2}*{MNIST} & \multirow{2}*{CIFAR10} & \multicolumn{3}{c}{ImageNet} \\
        \cmidrule(r){4-6}
        & & & Un-targeted & Targeted & Un-targeted Defense \\ \hline
        Sample size ($b$)& 20 & 20 & 20 & 20 & 20 \\
        Learning rate ($\eta$)& 0.3 & 2.0 & 5.0 & 3.0 & 5.0 \\ \hline
    \end{tabular}
    \label{tab:hyper_embedding}
\end{table}

\begin{table}[!h]
    \centering
    \setlength\tabcolsep{25.5pt}
    \caption{Hyperparameters for {\TGenAttack}. }
    \begin{tabular}{c c c} \hline
        &CIFAR10 Defense & ImageNet Defense  \\ \hline 
        Sample size ($b$)& 20 & 20 \\
        Learning rate ($\eta$)&  2.0 & 5.0 \\ 
        White-box iteration & 100 & 100 \\
        White-box margin($\kappa$) & 100 & 100 \\
        White-box learning rate & 1.0 & 2.0 \\\hline
    \end{tabular}
    \vspace{10cm}
    \label{tab:hyper_embeddingos}
\end{table}

\end{document}